\pdfoutput=1
\documentclass[11pt]{article}
\usepackage[final]{acl}
\usepackage{times}
\usepackage{latexsym}
\usepackage[T1]{fontenc}
\usepackage[utf8]{inputenc}
\usepackage{microtype}
\usepackage{inconsolata}
\usepackage{amsmath}
\usepackage{amssymb}
\usepackage{algorithm}
\usepackage{algpseudocode}
\usepackage{soul}
\usepackage{xspace}
\usepackage{booktabs}
\usepackage{multirow}
\usepackage{graphicx}
\usepackage{tabularx}
\usepackage{subcaption}
\usepackage{tcolorbox}

\newcommand{\paratitle}[1]{\vspace{1.5ex}\noindent\textbf{#1}}
\newcommand{\ie}{\emph{i.e.,}\xspace}

\newcommand{\eg}{\emph{e.g.,}\xspace}

\newcommand{\OURS}{\textbf{DAWN-ICL}\xspace}
\newcommand{\ignore}[1]{}

\title{\OURS: Strategic Planning of Problem-solving \\ Trajectories for Zero-Shot In-Context Learning}

\author{
    Xinyu Tang\textsuperscript{\thanks{\ \ Equal contribution.}},
    Xiaolei Wang\textsuperscript{\footnotemark[1]},
    Wayne Xin Zhao\textsuperscript{\thanks{\ \ Corresponding author.}},
    Ji-Rong Wen \\
    Gaoling School of Artificial Intelligence, Renmin University of China \\
    \texttt{txy20010310@163.com, wxl1999@foxmail.com, batmanfly@gmail.com}\\
}

\begin{document}
\maketitle

\begin{abstract}

Zero-shot in-context learning~(ZS-ICL) aims to conduct in-context learning~(ICL) without using human-annotated demonstrations.
Existing ZS-ICL methods either use large language models~(LLMs) to generate (input, label) pairs as \textit{pseudo-demonstrations} or leverage historical pseudo-demonstrations to help solve the current problem.
They assume that all problems are from the same task and traverse them in a random order.
However, in real-world scenarios, problems usually come from diverse tasks, and only a few belong to the same task.
The random traversing order may generate \textit{unreliable} pseudo-demonstrations and lead to \textit{error accumulation}.
To address this problem, we reformulate ZS-\textbf{ICL} as a \textit{planning} problem and propose a \textbf{D}emonstration-\textbf{AW}are Mo\textbf{N}te Carlo Tree Search~(MCTS) approach~(\OURS), which leverages MCTS to strategically plan the problem-solving trajectories for ZS-ICL.
In addition, to achieve effective and efficient $Q$ value estimation, we propose a demonstration-aware $Q$-value function and use it to enhance the \textit{selection} phase and accelerate the \textit{expansion} and \textit{simulation} phases in MCTS.
Extensive experiments demonstrate the effectiveness and efficiency of \OURS on in-domain and cross-domain scenarios, and it even outperforms ICL using human-annotated demonstrations.
The code is available at \url{https://github.com/txy77/MCTS4ZSICL}.

\end{abstract}
\section{Introduction}

\begin{figure}[t]
    \centering
    \includegraphics[width=\linewidth]{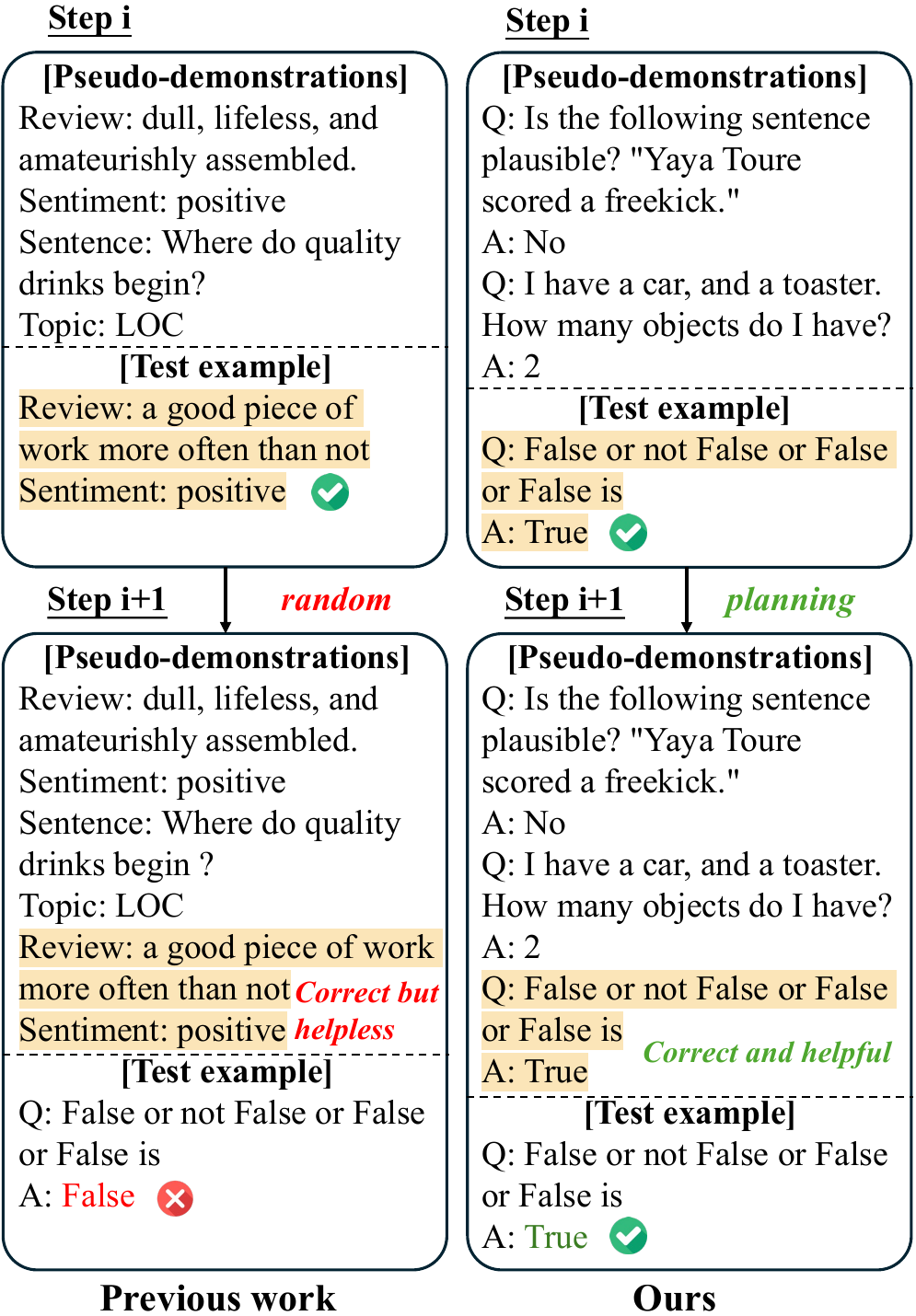}
    \caption{Comparison of our method with previous methods. Although both predictions at $i$-th step are correct, in previous work, the example is randomly selected and helpless. In contrast, in our method, the example is selected with planning and helpful.}
\label{fig:intro}
\end{figure}

In-context learning~(ICL)~\citep{NIPS-Brown-GPT3,arXiv-Dong-icl-survey, ICL-pretrain} represents a significant advancement in the capabilities of large language models~(LLMs).
It allows LLMs to rapidly adapt to new tasks without updating the parameters by adding only a few examples as demonstrations to the input.

Although no training is required, most ICL work assumes access to large-scale external resources (\eg training dataset~\cite{ACL-Liu-TopK, ICML-Ye-DPP} or relevant corpus~\cite{ACL-Tanwar-cross-domain-icl, ACL-Chatterjee-cross-domain-icl}) for demonstration selection, which is usually not available in real-world scenarios.
To eliminate this dependency, \textit{zero-shot in-context learning}~(ZS-ICL)~\cite{ACL-Lyu-ZICL, EMNLP-Chen-SelfICL} is proposed, which uses LLMs to generate (input, label) pairs as \textit{pseudo-demonstrations} for ICL.
However, since LLMs are limited in data synthesis~\cite{arXiv-Synthetic-Seddik,arXiv-Synthetic-Longpre,arXiv-Synthetic-Dohmatob}, the performance of ZS-ICL usually falls behind ICL with human-annotated demonstrations.
To remedy this deficiency, recent work~\cite{ACL-Su-DAIL} employs previously predicted examples as the source of demonstrations, eliminating the need for input synthesis and reusing the predicted labels.
This method assumes that test examples are from the same task and traverse them in a random order. 
However, in real-world scenarios, examples usually come from diverse tasks, and only a few belong to the same task.
The random traversing order may cause LLMs to generate \textit{unreliable} pseudo-demonstrations and lead to \textit{error accumulation}, as illustrated in Figure~\ref{fig:intro}.

To address the above problems, we aim to optimize the traversing order of examples in ZS-ICL and formulate it as a \textit{planning} problem.
To search the traversing order, we take inspiration from Monte Carlo Tree Search~(MCTS)~\cite{CG-Coulom-MCTS}, which can conduct a strategic tree search and strike a balance between exploration and exploitation.
In the algorithm, MCTS maintains a state-action value function $Q(s, a)$ to estimate the expected future reward of taking action $a$ in state $s$, which is updated by the simulation and back-propagation step in each iteration.
However, for ZS-ICL, such an updating method is too costly to achieve accurate estimation since the state space is very large ($n!$ for $n$ examples), and each state requires the LLM to perform one inference for reward calculation.

To this end, in this paper, we propose a novel \textbf{D}emonstration-\textbf{AW}are Mo\textbf{N}te Carlo Tree Search for ZS-\textbf{ICL}, namely \OURS.
Our core idea is to leverage MCTS for planning the problem-solving trajectories in ZS-ICL.
To achieve effective and efficient $Q$ value estimation in MCTS, we propose to integrate the information of the pseudo-demonstration set into the $Q$-value function.
With this demonstration-aware $Q$-value function, we can enhance the \textit{selection} phase and accelerate the \textit{expansion} and \textit{simulation} phases of MCTS for more effective and efficient search.
Furthermore, we design a calibration-enhanced aggregation method to derive the final prediction from MCTS, which aggregates results from multiple iterations and debiases the prediction with pre-trained priors.
To validate the effectiveness of our approach, we conduct experiments on the in-domain and cross-domain scenarios of BBH and MMLU across various LLMs.
The experimental results show that \OURS consistently surpasses the best ZS-ICL baseline method and even outperforms ICL using human-annotated demonstrations.

Our contributions can be summarized as follows:

$\bullet$ To the best of our knowledge, we are the first to formalize ZS-ICL as a planning problem, which is closer to real-world scenarios.

$\bullet$ We propose a demonstration-aware MCTS for ZS-ICL to achieve a more effective and efficient search for the problem-solving trajectories.

$\bullet$ Extensive experiments demonstrate the effectiveness of our approach on in-domain and cross-domain scenarios, and it even outperforms ICL using human-annotated demonstrations.
\section{Related Work}


\paratitle{Zero-shot In-context Learning.}
Zero-shot in-context learning~(ZS-ICL)~\cite{ACL-Lyu-ZICL, EMNLP-Chen-SelfICL, ACL-Su-DAIL} aims to conduct in-context learning~(ICL) using model-generated pseudo-demonstrations.
Most ZS-ICL work separately generates pseudo-demonstrations for each example~\cite{ACL-Lyu-ZICL, EMNLP-Chen-SelfICL}.
There is also some work that employs previously predicted examples as demonstrations and stores them in memory for future usage~\cite{ACL-Su-DAIL}.
However, these methods traverse examples in a random order, which may lead to error accumulation.
In this paper, we reformulate ZS-ICL as a planning problem and search for the optimal problem-solving order.

\paratitle{Enhancing LLMs with Planning.}
Recent advancements in enhancing LLMs through planning have shown promising results~\cite{NIPS-Yao-ToT, arXiv-Hao-RAP, ICML-Wan-AlphaZero, arXiv-Wang-Q*}.
They often engage in deliberate reasoning processes by utilizing strategic planning algorithms like MCTS~\cite{CG-Coulom-MCTS} to explore intermediate steps.
{In this work, we aim to better utilize the in-context learning ability of LLMs by performing a novel demonstration-aware MCTS for ZS-ICL over the demonstration space.}

\label{sec-related_work}
\section{Methodology}
\label{sec:method}

\begin{figure*}[t]
    \centering
    \includegraphics[width=\textwidth]{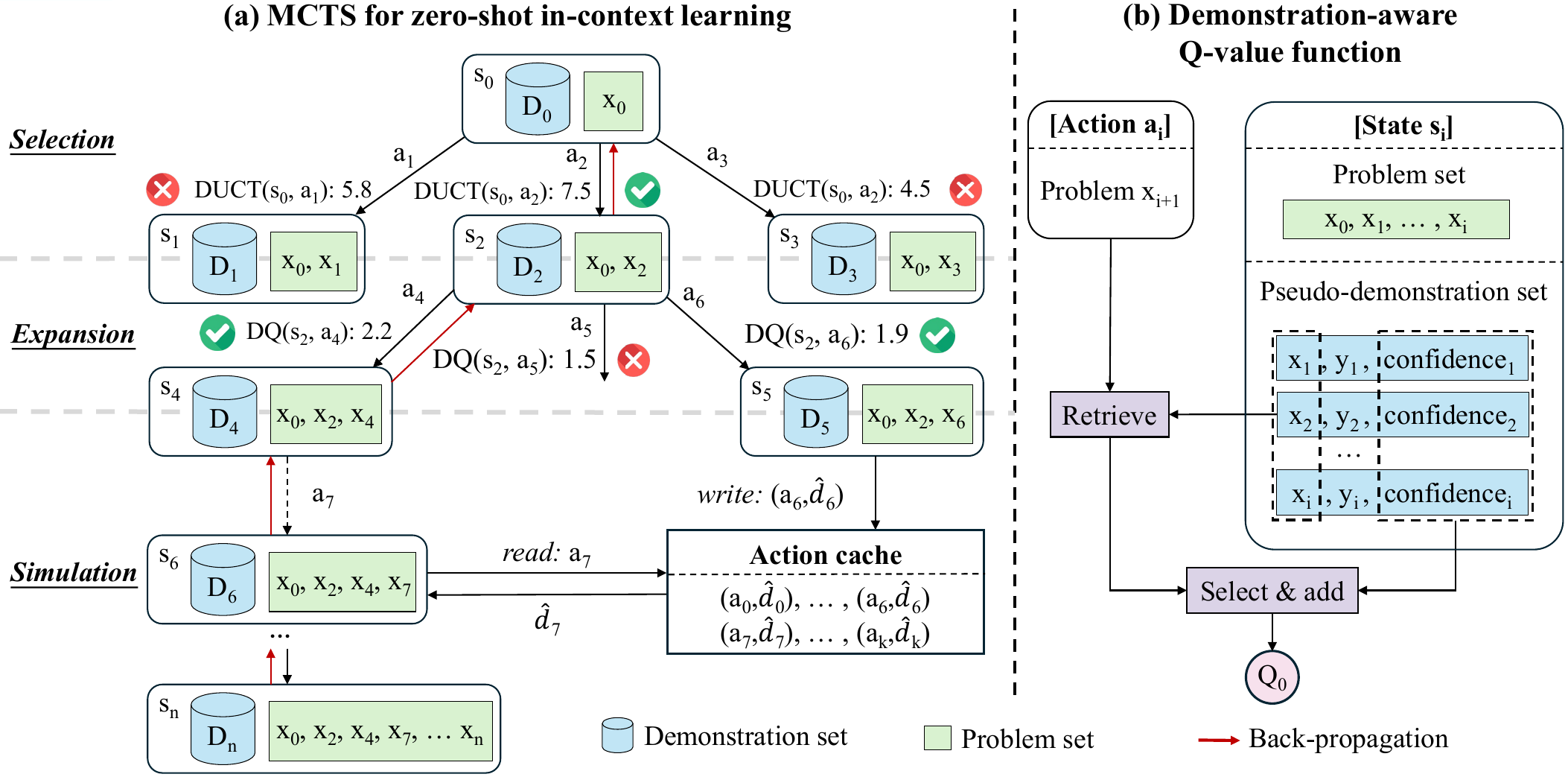}
    \caption{The overview of \OURS. (a) An illustration of the four phases in MCTS. We select nodes using our proposed DUCT (Eq.~\ref{eq:selection}), perform expansion using our proposed DQ function (Eq.~\ref{eq:Q-func}), accelerate simulation with an action cache supported by the DQ function, and finally back-propagate the rewards. (b) We improve the $Q$-value function with pseudo-demonstration information. We retrieve $k$ pseudo-demonstrations and add the score of confidence and similarity as the initial value of the $Q$ function.}
\label{fig:main}
\end{figure*}

In this section, we present our demonstration-aware MCTS for ZS-ICL, namely \OURS.
We first give an overview of our approach, then discuss how to integrate the information of demonstrations into the $Q$-value function, and finally present the planning approach to ZS-ICL.
The overall architecture of \OURS is illustrated in Figure~\ref{fig:main}.

\subsection{Overview of Our Approach}
\label{subsec:overview}

\paratitle{Problem formulation.}
Zero-shot in-context learning~(ZS-ICL) enables task adaptation of LLMs at test time without using human-labeled examples as the demonstration.
Formally, given an LLM $\mathcal{M}$ and a set of $n$ test examples $\mathcal{E} = \{(x_k, y_k)\}_{k=1}^n$, at $i$-th step, an example $x_i$ is first selected from $\mathcal{E}$, and then the pseudo-demonstration $d_i = f_c (x_i, \mathcal{D}_{i-1})$ for $x_i$ is constructed based on the existing pseudo-demonstration set $\mathcal{D}_{i-1}$.
Based on this, the LLM makes the prediction $\hat{y}_i = f_p (x_i, d_i ; \mathcal{M})$ and obtains a new pseudo-demonstration $\hat{d}_i = (x_i, \hat{y}_i)$.
Then, the pseudo-demonstration set is updated as $\mathcal{D}_{i} = f_u (\mathcal{D}_{i-1}, \hat{d}_i)$.
The above process is repeated until all the problems in $\mathcal{E}$ are solved.

\paratitle{The general planning framework.}
In existing work~\cite{ACL-Su-DAIL}, the test examples are assumed to belong to the same task, and the example to solve at each step ($x_i$) is usually randomly selected.
However, in real-world scenarios, test examples come from diverse tasks, and only a few belong to the same task, limiting the performance.
To solve this, one feasible way is to optimize the traversing order of test examples.
This is because ICL is sensitive to the selection of demonstrations~\cite{ACL-Liu-TopK}.
Hence, the order of traversing is important for effectively leveraging historical examples as pseudo-demonstrations to help solve the current one.
To this end, our idea is to formalize ZS-ICL as a planning problem using a Markov Decision Process~(MDP), represented by the tuple $(\mathcal{S},\mathcal{A},T,r)$.
In this planning framework, we define the state $s_i \in \mathcal{S}$ as the set of test examples that have been solved at $i$-th step, along with the pseudo-demonstration set $\mathcal{D}_i$.
The action $a_i \in \mathcal{A}$ is to select the next problem $x_{i+1}$ to solve.
The transition function $T$ from the current state $s_i$ to the next state $s_{i+1}$ first performs the pseudo-demonstration construction function $f_c$, then the prediction function $f_p$ to solve $x_{i+1}$, and finally the pseudo-demonstration set updating function $f_u$ to obtain the new state $s_{i+1}$.
The reward function $r_i = r(s_i, a_i)$ measures the quality of the action $a_i$ applied to the state $s_i$.
Since ground-truth labels are not available at test time, we take the confidence of model prediction in $f_p$ as the reward, which has been shown to be aligned with the model performance~\cite{ICLR-Xiong-Confidence,arXiv-Zhang-Confidence}.

\paratitle{Monte Carlo Tree Search for zero-shot ICL.}
Our approach is inspired by the powerful planning algorithm Monte Carlo Tree Search~(MCTS), which can be used to strategically conduct tree search for problem-solving trajectories in ZS-ICL and strike a balance between exploration and exploitation to find high-reward trajectories.
To perform an effective search, MCTS maintains a state-action value function $Q: \mathcal{S} \times \mathcal{A} \to \mathbb{R}$, where $Q(s, a)$ estimates the expected future reward of taking action $a$ in state $s$.
For ZS-ICL, the quality of the pseudo-demonstration set in the current state is also an important signal for $Q$ values since ICL is known to be sensitive to demonstrations~\cite{EMNLP-Yoo-GroundTruthMatters,ACL-Liu-TopK}.
Taking this into consideration, we design specific metrics to integrate this information and propose a \textit{demonstration-aware} $Q$-value function (Section~\ref{subsec: q-value}).
Based on this, we conduct MCTS for ZS-ICL (Section~\ref{subsec:SPICL}), where the proposed $Q$-value function is used to enhance the \textit{selection} phase and accelerate the \textit{expansion} and \textit{simulation} phases.
Furthermore, we design a calibration-enhanced aggregation method to derive the final prediction from MCTS, which aggregates results from multiple iterations and debiases the prediction with pre-trained priors.
In what follows, we introduce these two parts in detail.

\subsection{Demonstration-aware $Q$-value Function}
\label{subsec: q-value}

The MCTS algorithm maintains a state-action value function $Q(s, a)$ to estimate the expected reward of taking the action $a$ in the state $s$.
Originally, this $Q$-value function is updated by simulating the future states (\ie simulation) and aggregating their rewards (\ie back-propagation).
However, for ZS-ICL, such an updating method is too costly to achieve accurate estimation since the state space is very large, and the reward calculation of each state requires the LLM to perform one inference.

To perform effective $Q$ value estimation, we propose to leverage the contextual information of the current state and action to initialize the $Q$ value.
The performance of ICL is known to be highly dependent on the selection of demonstrations.
Inspired by this, we propose to initialize the $Q$ value by evaluating the quality of the pseudo-demonstration set $D_i$ in the current state $s_i$ with respect to the problem $x_{i+1}$ (\ie action $a_i$).
Specifically, we first retrieve $k_d$ demonstrations that are most semantically similar to the problem, and then evaluate their quality by aggregating their confidence and similarity scores.
The demonstration-aware $Q$-value function DQ can be represented as follows:
\begin{align}
    \text{DQ}(s, a)  &= Q_0(s, a) + w_Q \cdot Q(s, a), \label{eq:Q-func} \\
    Q_0(s, a) &= \frac{1}{k_d} \sum_{i=1}^{k_d} \left(C(d_i) + S(d_i, x_{i+1})\right),
\end{align}
where $d_i$ is the retrieved demonstration, {$C(d_i)$ is the confidence score of the demonstration from the prediction function $f_p$, $S(d_i, x_{i+1})$ is the similarity score between the demonstration and the problem $x_{i+1}$ chosen by the action measured with the BGE model~\cite{SIGIR-Xiao-BGE}, and $w_Q$ is a constant to balance the initial value ($Q_0(s, a)$) and updated value ($Q(s, a)$).

With this demonstration-aware $Q$-value function, the estimation of $Q$ values can be more accurate with a limited computational budget.
Based on this, we conduct MCTS for ZS-ICL, where the demonstration-aware $Q$-value function is used to enhance the \textit{selection} step and accelerate the \textit{expansion} and \textit{simulation} step.
In the next section, we will introduce this in detail.

\subsection{Strategic Planning for Zero-shot In-context Learning}
\label{subsec:SPICL}

In this section, we first introduce the demonstration-aware MCTS to plan problem-solving trajectories for ZS-ICL, then detail the calibration-enhanced aggregation method to derive the final prediction from the searched trajectories.

\subsubsection{Demonstration-aware MCTS}

The reformulation of ZS-ICL as a planning problem (Section~\ref{subsec:overview}) enables us to leverage principle planning algorithms, notably the Monte Carlo Tree Search~(MCTS).
Specifically, MCTS iteratively constructs a tree for search, where each node represents a state and each edge represents an action and the transition from the current state to the next one by applying the action.
Each iteration consists of four phases: selection, expansion, simulation, and back-propagation.
For ZS-ICL, we randomly select a problem with the empty pseudo-demonstration set as the initial state and execute the algorithm until a predefined number of iterations is reached.
In each iteration, we use the proposed demonstration-aware $Q$-value function to enhance the \textit{selection} phase and accelerate the \textit{expansion} and \textit{simulation} phases.
Next, we detail the four phases of MCTS for planning problem-solving trajectories in ZS-ICL.
The pseudo-code is presented in Algorithm~\ref{alg:mcts}.

\paratitle{Selection.}
The first phase of MCTS is to select the most promising part of the existing tree for expansion.
A well-known selection strategy is the Upper Confidence bounds applied to Trees~(UCT) algorithm~\cite{ECML-Kocsis-UCT}, which can effectively balance exploration (less visited times) and exploitation (high $Q$ values).
However, as mentioned in Section~\ref{subsec: q-value}, the estimation of $Q$ values is too costly to be accurate in ZS-ICL.
Therefore, we propose to integrate our DQ function (Eq.~\ref{eq:Q-func}) into UCT.
This demonstration-aware UCT function DUCT can be represented as follows:
\begin{equation}
\label{eq:selection}
    \text{DUCT}(s, a) = \text{DQ}(s, a) + w_a \sqrt{\frac{\ln N(s)}{N (\text{c}(s, a))}},
\end{equation}
where $N(s)$ is the number of times node $s$ has been visited, $\text{c}(s, a)$ is the child node for $s$ after applying action $a$, and $w_a$ is a constant to balance exploration ($U(s, a)$) and exploitation ($\text{DQ}(s, a)$).
We start from the root node~(\ie initial state $s_0$) and repeatedly select a child node with the maximum DUCT value until reaching a leaf node.

\paratitle{Expansion.}
This phase expands the tree by generating child nodes for the leaf node selected above.
Since the action space of a state (\ie remaining problems to solve) can be large, we use our proposed DQ function for efficient action selection.
Specifically, we first calculate the value of each action using Eq.~\ref{eq:Q-func}, and then choose the top-$k_a$ actions with the highest values for expansion.
For each selected action $a_i$, we need to predict the corresponding state $s_{i+1}$, which is the role of the transition function $T$ described in Section~\ref{subsec:overview}.
In $T$, the first step is to construct the pseudo-demonstrations for the problem $x_{i+1}$ selected by the action (\ie function $f_c$).
To make the pseudo-demonstrations relevant and diverse, we first retrieve $k$ most semantically similar ones with the problem $x_{i+1}$ from the pseudo-demonstration set to increase the relevance, and then randomly select samples with different pseudo-labels from them to enhance the diversity.
With the pseudo-demonstrations $d_{i+1}$, the second step in $T$ is to predict the label for the problem $x_{i+1}$ using the LLM (\ie function $f_p$).
Here, we use greedy decoding to generate the prediction.
Finally, the predicted label $\hat{y}_{i+1}$ paired with the problem $x_{i+1}$ is added to the pseudo-demonstration set (\ie function $f_u$), and the new state $s_{i+1}$ is obtained.
For the expanded nodes, we choose the one with the largest reward for the next simulation phase.

\paratitle{Simulation.}
This phase simulates the future trajectories for the node selected from the previous expansion step.
The simulation process typically involves a roll-out policy to reach the terminal state and calculate the future rewards.
For simplicity, we follow the same procedure as the expansion phase, \ie selecting $k_a$ candidate actions with the highest DQ values and picking the one with the largest reward.
To further accelerate the simulation process, we propose a cache mechanism based on the DQ function.
Specifically, we maintain the maximum DQ value for each action $a_i$ as $\text{DQ}_{\max}^{(i)}$ and record the corresponding pseudo-demonstration $\hat{d}_i = (x_{i+1}, \hat{y}_{i+1})$.
If $\text{DQ}_{\max}^{(i)}$ breaks through the threshold $\epsilon$, we add the ($a_i$, $\hat{d}_i$) pair into the cache.
In the simulation process, if we take an action that exists in the cache, the pseudo-demonstration is read from the cache, and we skip the transition function $T$ to directly obtain the new state.

\paratitle{Back-propagation.}
This phase is executed when we reach a terminal node.
We back-propagate the rewards along the path from the terminal node to the root node by updating the $Q$-value function.
Specifically, we update $Q(s_i, a_i)$ by calculating the mean rewards in all the future trajectories starting from $s_i$.

\subsubsection{Calibration-Enhanced Aggregation}
\label{subsubsec:aggregation}

The above MCTS algorithm could produce multiple trajectories and predictions for each test example through multiple iterations.
In this part, we introduce a calibration-enhanced aggregation method to produce the final answer while debiasing the answer with pre-trained priors.

\paratitle{Aggregation.}
Considering that in ICL, the unique correct answer can be derived from multiple different demonstrations, we collect the predictions from each iteration of MCTS to make the final prediction.
Specifically, we calculate the average probabilities for each label and select the one with the highest probability as the final answer, which can be represented as follows:
\begin{equation}
    y^* = \mathop{\arg\max}_{y_i \in \mathcal{Y}} \frac{1}{N_i} \sum_{j=1}^{N_i} \text{Pr}(y_i | x, d_j ; \mathcal{M}),
\end{equation}
where $\mathcal{Y}$ is the label space, $N_i$ is the number of predictions for label $y_i$, and $\text{Pr}(y_i | x, d_j ; \mathcal{M})$ is the probability for $y_i$ given by the LLM $\mathcal{M}$ with problem $x$ and pseudo-demonstration $d_j$ as the input.

\paratitle{Calibration.}
LLMs are known to suffer from common token bias~\cite{ICML-Zhao-calibration}, which means they are biased towards tokens common in their pre-training data.
To debias the prediction of LLMs, we adopt a calibration strategy based on prior probability.
Specifically, we first obtain the prior probability of each label by calculating its average probability predicted by the LLM across all the test examples.
Then, we derive the calibrated probability of each prediction by dividing the prior probability.
We can integrate this strategy with aggregation, which is represented as follows:
\begin{gather}
    y^* = \mathop{\arg\max}_{y_i \in \mathcal{Y}} \frac{1}{N_i} \sum_{j=1}^{N_i} \frac{\text{Pr}(y_i | x, d_j ; \mathcal{M})}{\text{Pr}(y_i | \mathcal{M})}, \\
    \text{Pr}(y_i | \mathcal{M}) = \frac{1}{|\mathcal{E}|} \sum_{j=1}^{|\mathcal{E}|} \text{Pr}(y_i | x_j ; \mathcal{M}),
\end{gather}
where $\text{Pr}(y_i | \mathcal{M})$ is the prior probability of the LLM $\mathcal{M}$ for $y_i$ and $\text{Pr}(y_i | x_j ; \mathcal{M})$ is the zero-shot probability for $y_i$ given by the LLM $\mathcal{M}$ with only the problem $x_j$ as the input.
\section{Experiments}

In this section, we first set up the experiments, then report the results and conduct a detailed analysis.

\subsection{Experimental Setup}

\paratitle{Datasets.}
To evaluate the effectiveness of our method, following~\citet{ACL-Su-DAIL}, we conduct experiments on the BIG-Bench Hard~(BBH)~\cite{ACL-Suzgun-BBH-Dataset} and Massive Multitask Language Understanding~(MMLU)~\cite{ICLR-Hendrycks-MMLU-Dataset} benchmarks.
Specifically, we consider two scenarios: in-domain and cross-domain.
For the in-domain scenario, we evaluate each task of BBH and MMLU separately.
For the cross-domain scenario, we randomly select 8 samples from each task of BBH to construct a dataset called BBH-mini.

\paratitle{Baselines.}
To facilitate a systematic comparison, we select several representative methods:

$\bullet$ \textbf{\underline{Zero-shot}}:
The model directly makes predictions without any demonstration.

$\bullet$ \textbf{\underline{Few-shot}}~\cite{NIPS-Brown-GPT3}:
The model makes predictions with human-annotated demonstrations.
It is not entirely fair to compare it with other methods, as it uses external information.

$\bullet$ \textbf{\underline{Self-ICL}}~\cite{EMNLP-Chen-SelfICL}:
The model makes predictions with self-generated pseudo-demonstrations.

$\bullet$ \textbf{\underline{DAIL}}~\cite{ACL-Su-DAIL}:
The model makes predictions with pseudo-demonstrations retrieved from previously predicted examples.

\paratitle{Models.}
We use representative open-source LLMs (\ie Llama3.1-8B~\cite{arXiv-Dubey-llama3.1}, Qwen2.5-7B~\cite{qwen2}, and Mistral-7B-v0.3~\cite{arXiv-Jiang-Mistral} and a close-source LLM (\ie GPT-4o-mini~\cite{OpenAI-Blog-GPT-4o-mini}) for experiment.

\paratitle{Implementation Details.}
In MCTS, we set the number of iterations as {5}.
For the $Q'$ function, we set {$k_d$ as 30} and $w_Q$ as 1.
For the selection phase, we set $w_a$ in Eq.~\ref{eq:selection} as 5.
For the expansion phase, we set {$k_a$ as 3}.
For the simulation phase, we set $\epsilon$ as 1.5.
Following \citet{ACL-Su-DAIL}, we set the number of demonstrations in $f_p$ to 3 for BBH and 4 for MMLU.
Notably, we can not obtain the logits of GPT-4o-mini, so we do not use calibration for it.
To test the full potential of our approach, we also consider removing the cache mechanism in the simulation phase, named ``DAWN-ICL w/o cache''.

\subsection{Experimental Results}
\begin{table*}[t]
\centering
\resizebox{\textwidth}{!}{
\begin{tabular}{cl|ccc|ccccc}
\toprule
\multicolumn{2}{c|}{\multirow{3.5}{*}{\textbf{Task}}} & \multicolumn{3}{c|}{\textbf{BBH}} & \multicolumn{5}{c}{\textbf{MMLU}} \\
\cmidrule{3-10}
& & \begin{tabular}[c]{@{}c@{}}\textbf{Binary}\\ \textbf{Choice}\end{tabular} & \begin{tabular}[c]{@{}c@{}}\textbf{Multiple}\\ \textbf{Choice}\end{tabular} & \textbf{Average} & \textbf{STEM} & \begin{tabular}[c]{@{}c@{}}\textbf{Human}\\ \textbf{anities}\end{tabular} & \begin{tabular}[c]{@{}c@{}}\textbf{Social}\\ \textbf{Sciences}\end{tabular} & \textbf{Other} & \textbf{Average} \\
\midrule
\multirow{6}{*}{\textbf{Llama3.1-8B}}
& Zero-shot             & 52.26         & 38.81           & 42.32  & 48.91 & 52.22      & 68.28         & 65.82 & 58.00  \\
& Few-shot              & 53.93         & 42.42           & 45.42  & 51.92 & 56.77      & 73.94         & 69.13 & 62.18  \\
& Self-ICL              & 50.87         & 31.64           & 36.65  & 43.51 & 47.57      & 61.94         & 58.77 & 52.29  \\
& DAIL                  & 52.82         & 39.08           & 42.66  & 52.65 & 56.30      & 75.14         & 70.04 & 62.65  \\
\cmidrule{2-10}
& \OURS             & \underline{60.26}  & \underline{43.69}  & \underline{48.01}   & \underline{54.04}  & \underline{58.11} & \underline{75.24} & \underline{70.94}  & \underline{63.79}  \\
& \OURS w/o cache   & \textbf{61.86} & \textbf{43.86}  & \textbf{48.56}      & \textbf{54.65} & \textbf{58.94} & \textbf{76.34} & \textbf{71.58} & \textbf{64.58}      \\
\midrule
\multirow{6}{*}{\textbf{Qwen2.5-7B}}
& Zero-shot             & 59.71         & 46.56           & 49.99   & 64.03 & 59.64      & 80.18         & 74.90 & 68.50  \\
& Few-shot              & 55.74         & \textbf{50.76}  & 52.06   & 67.68 & 64.31      & 82.03         & 75.99 & 71.54  \\
& Self-ICL              & 52.75         & 45.83           & 47.63   & 62.92 & 56.77      & 75.98         & 71.29 & 65.57  \\
& DAIL                  & 55.74         & 47.25           & 49.46   & 67.71 & 64.59      & 82.74         & \underline{76.29} & 71.86  \\
\cmidrule{2-10}
& \OURS             & \underline{64.51} & 49.21           & \underline{53.20}  & \textbf{68.41}   & \underline{64.91}  & \underline{83.26}          & 75.47          & \underline{72.06}          \\
& \OURS  w/o cache  & \textbf{65.90} & \underline{50.17} & \textbf{54.27}     & \underline{68.03} & \textbf{65.27} & \textbf{83.82} & \textbf{76.44} & \textbf{72.43}     \\
\midrule
\multirow{6}{*}{\textbf{Mistral-7B}}
& Zero-shot             & 53.79         & 33.46           & 38.76  & 46.69          & 52.75      & 66.17         & 64.85 & 57.01  \\
& Few-shot              & 60.26         & 40.03           & 45.31  & \underline{51.74} & 54.98      & 71.40         & 68.10 & 60.70  \\
& Self-ICL              & 56.44         & 33.51           & 39.48  & 40.63          & 48.20      & 58.21         & 58.80 & 51.04  \\
& DAIL                  & 58.39         & 36.43           & 42.15  & 50.62          & 54.54      & 72.44         & 68.59 & 60.69  \\
\cmidrule{2-10}
& \OURS             & \underline{61.10} & \underline{41.80} & \underline{46.83} & 51.51          & \underline{57.32} & \underline{72.77}   & \textbf{68.97}          & \underline{61.98}  \\
& \OURS w/o cache   & \textbf{62.14} & \textbf{42.24}  & \textbf{47.43}      & \textbf{52.14}    & \textbf{58.51} & \textbf{73.06} & \underline{68.78} & \textbf{62.54}      \\
\midrule
\multirow{6}{*}{\textbf{GPT-4o-mini}}
& Zero-shot            & 52.26          & 37.43           & 41.30  & 40.37 & 52.77      & 59.05         & 56.90          & 52.28  \\
& Few-shot             & \textbf{65.97} & 49.56           & 53.84  & 45.26 & 64.34      & 71.17         & \underline{69.07}        & 62.60  \\
& Self-ICL             & 63.47          & 47.91           & 51.97  & 48.37 & 59.17      & \underline{71.56} & 65.59          & 60.88  \\
& DAIL                 & 64.44          & 50.00           & 53.77  & 44.08 & 61.91      & 61.46         & 59.25          & 57.22  \\
\cmidrule{2-10}
& \OURS                & 62.98          & \underline{54.10}  & \underline{56.41} & \underline{51.95} & \underline{65.65} & 70.75          & 68.20          & \underline{64.26}      \\
& \OURS  w/o cache     & \underline{64.93} & \textbf{54.25}  & \textbf{57.03}    & \textbf{53.38} & \textbf{66.18} & \textbf{73.12} & \textbf{69.18} & \textbf{65.62}   \\
\bottomrule
\end{tabular}
}
\caption{Performance comparison across various LLMs on the in-domain scenario using BBH and MMLU. The best method in each group is marked in \textbf{bold}, and the second-best method is marked with an \underline{underline}.}
\label{tab:main_exp}
\end{table*}

\begin{table}[t]
\centering
\resizebox{\linewidth}{!}{
    \begin{tabular}{l|cccc}
    \toprule
    \textbf{Model} & \begin{tabular}[c]{@{}c@{}}\textbf{Llama3.1}\\ \textbf{-8B}\end{tabular} & \begin{tabular}[c]{@{}c@{}}\textbf{Qwen2.5}\\ \textbf{-7B}\end{tabular} & \begin{tabular}[c]{@{}c@{}}\textbf{Mistral}\\ \textbf{-7B}\end{tabular} & \begin{tabular}[c]{@{}c@{}}\textbf{GPT-4o}\\ \textbf{-mini}\end{tabular} \\
    \midrule
    Zero-shot      & 36.41 & 48.37 & 36.41 & 37.50 \\
    Few-shot       & 40.76 & 49.46 & 39.67 & 54.35 \\
    Self-ICL       & 40.22 & 47.28 & 35.87 & 51.09 \\
    DAIL~(random)    & 35.33 & 44.02 & 39.13 & 50.00 \\
    DAIL~(sequential)    & 39.13 & 47.28 & 40.22 & 54.89 \\
    \midrule
    \OURS  & \underline{43.48}       & \textbf{51.09}    & \textbf{47.83}    & \underline{55.43}  \\
    \OURS w/o cache & \textbf{44.57} & \underline{50.54} & \underline{44.57} & \textbf{59.24}     \\
    \bottomrule
    \end{tabular}
}
\caption{Performance comparison across various LLMs on the cross-domain scenario using BBH-mini. The best method in each group is marked in \textbf{bold}, and the second-best method is marked with an \underline{underline}.}
\label{tab:mini_exp}
\end{table}

\paratitle{In-domain scenario.}
Table~\ref{tab:main_exp} presents the results across various LLMs on the in-domain scenario.
As we can see, Self-ICL performs poorly and even worse than zero-shot prompting for some LLMs.
The main reason is that LLMs are limited in data synthesis~\cite{arXiv-Synthetic-Seddik} and may even have insufficient domain knowledge for specific tasks.
Thus, they struggle to generate high-quality pseudo-demonstrations.
In contrast, DAIL shows decent improvements and consistently outperforms zero-shot prompting.
DAIL employs previously predicted examples as the source of demonstrations, eliminating the need for input synthesis to improve the quality of pseudo-demonstrations.
However, DAIL still underperforms few-shot prompting sometimes, as it randomly selects examples at each step, and the historical examples may not be beneficial for the current one. 
Finally, \OURS surpasses all the ZS-ICL baselines by a large margin and even consistently outperforms few-shot prompting.
We reformulate ZS-ICL as a planning problem and use MCTS to search for the optimal traversing order.
Thus, the historical examples can better help the current one, as they are selected based on $Q'$ values rather than chosen at random.
{In addition, we observe that integrating a cache mechanism has minimal impact on performance but can significantly speed up the expansion and simulation processes of MCTS.
More detailed analysis of search efficiency are presented in Section~\ref{subsubsec:search_strategy}.}

\paratitle{Cross-domain scenario.}
For the cross-domain scenario, the results are shown in Table~\ref{tab:mini_exp}.
Similar to the in-domain scenario, Self-ICL and DAIL~(random) with the random traversing order perform poorly.
In contrast, if DAIL sequentially deals with each task (\ie DAIL~(sequential)), the performance improves greatly, which means that traversing order is important, especially in the cross-domain scenario.
Finally, \OURS achieves the best performance and even surpasses the few-shot prompting.
It demonstrates that our approach is generally applicable to various scenarios in ZS-ICL.

\subsection{Detailed Analysis}

In this part, we construct a detailed analysis of the effectiveness of our approach.

\subsubsection{Ablation Study}

\begin{table}[t]
\centering
\resizebox{\linewidth}{!}{
\begin{tabular}{l|cc|cc}
\toprule
\textbf{Task} & \multicolumn{2}{c|}{\textbf{BBH}} & \multicolumn{2}{c}{\textbf{MMLU}} \\
\midrule
\textbf{Model} & \begin{tabular}[c]{@{}c@{}}\textbf{Llama3.1}\\ \textbf{-8B}\end{tabular} & \begin{tabular}[c]{@{}c@{}}\textbf{Qwen2.5}\\ \textbf{-7B}\end{tabular} & \begin{tabular}[c]{@{}c@{}}\textbf{Llama3.1}\\ \textbf{-8B}\end{tabular} & \begin{tabular}[c]{@{}c@{}}\textbf{Qwen2.5}\\ \textbf{-7B}\end{tabular} \\
\midrule
\OURS               & 48.01 & 53.20 & 63.79 & 72.06 \\
\midrule
w/o $Q_0$ function  & 47.12 & 52.53 & 63.20 & 71.88 \\
w/o aggregation     & 45.94 & 53.08 & 63.15 & 71.98 \\
w/o calibration     & 44.96 & 51.12 & 63.11 & 71.91 \\
\bottomrule
\end{tabular}
}
\caption{Ablation study on BBH and MMLU.}
\label{tab:ablation}
\end{table}

Our approach incorporates several important components to improve search quality and performance.
To validate the effectiveness of each component, we conduct the ablation study by removing demonstration-aware initial $Q_0$ value function, aggregation, and calibration strategies, respectively.
The results are shown in Table~\ref{tab:ablation}.
We can see that removing any component would lead to performance degradation.
It indicates that all the components in our model are helpful.
Among them, performance decreases the most after removing the calibration strategy.
This indicates the importance of calibration in our approach since it can effectively mitigate the inherent biases of LLMs.

\subsubsection{The Impact of Search Strategy}

\begin{table}[t]
\centering
\resizebox{\linewidth}{!}{
\begin{tabular}{l|rrrrr}
\toprule
\textbf{Task} & {MC} & {Greedy} & {Beam} & MCTS & \begin{tabular}[c]{@{}c@{}}MCTS\\w/o cache\end{tabular}  \\
\midrule
Boolean                         & 75.20         & 79.20         & 82.00         & 80.40         & \textbf{83.20} \\
Formal                          & 50.40         & 52.40         & 52.40         & 52.00         & \textbf{54.00} \\
Geometric                       & 28.80         & 33.60         & 40.80         & \textbf{43.60} & 42.40         \\
Hyperbaton                      & 70.40         & 71.60         & 78.00         & 78.80         & \textbf{80.40} \\
Movie                           & 72.40         & 78.80         & 81.60         & 76.00         & \textbf{83.60} \\
Reasoning                       & 37.60         & 41.60         & 45.60         & \textbf{50.00} &       {48.40} \\
Snarks                          & 57.30         & 57.30         & 59.55         & 58.43         & \textbf{61.80} \\
\midrule
\begin{tabular}[c]{@{}c@{}}Accuracy\end{tabular}      & 56.01         & 59.21         & 62.85         & 62.75         & \textbf{64.83} \\
\begin{tabular}[c]{@{}c@{}}Time\end{tabular}          & $\times$1.0             & $\times$1.0  & $\times$14.8  &  $\times$4.8 & $\times$14.8   \\
\hline
\end{tabular}
}
\caption{Average performance and inference time of different search methods on the selected tasks of BBH.}
\label{tab:search_strategy}
\end{table}

\label{subsubsec:search_strategy}

To systematically investigate the MCTS-based planning method in our approach, we conduct a comprehensive ablation study by comparing it with several alternative search methods on Llama3.1-8B, including a single Monte Carlo~(MC) search, greedy search, and beam search.
To ensure a fair comparison, we replace MCTS with other search methods while keeping other factors unchanged.
Specifically, MC search is a directionless method, which randomly selects one action at each step.
Greedy search selects the action with the highest DUCT score at each step.
Both methods perform a single pass through the problem space, meaning that LLMs generate one answer for each sample, which we use as the final answer.
Beam search maintains multiple promising paths by selecting the top beam-size paths with the maximum mean DUCT values at each step.
To keep a similar exploration space with MCTS, we set the expansion number for each node~(beam size) to 3 and keep the best 5 nodes for the next iteration to obtain the same number of trajectories.
In addition, we use the same calibration-enhanced aggregation method in Section~\ref{subsubsec:aggregation} to obtain the prediction results.

The results are shown in Table~\ref{tab:search_strategy}.
Greedy search is better than MC, which suggests the importance of a structured search strategy.
Beam search further improves the performance by maintaining multiple paths and exploring them in parallel.
However, these methods cannot strategically explore the problem space since they cannot look ahead and backtrack.
Thus, they are short in either performance or efficiency.
In contrast, our demonstration-aware MCTS can perform better with less inference budget.
Our approach explores the problem space with DCUT and efficiently looks ahead through memory-augmented simulation, making it more powerful for the complex search space in ZS-ICL.

\subsubsection{Exploration Efficiency Analysis}
\begin{figure}[t]
    \centering
    \begin{subfigure}[b]{0.49\linewidth}
        \centering
        \includegraphics[width=\textwidth]{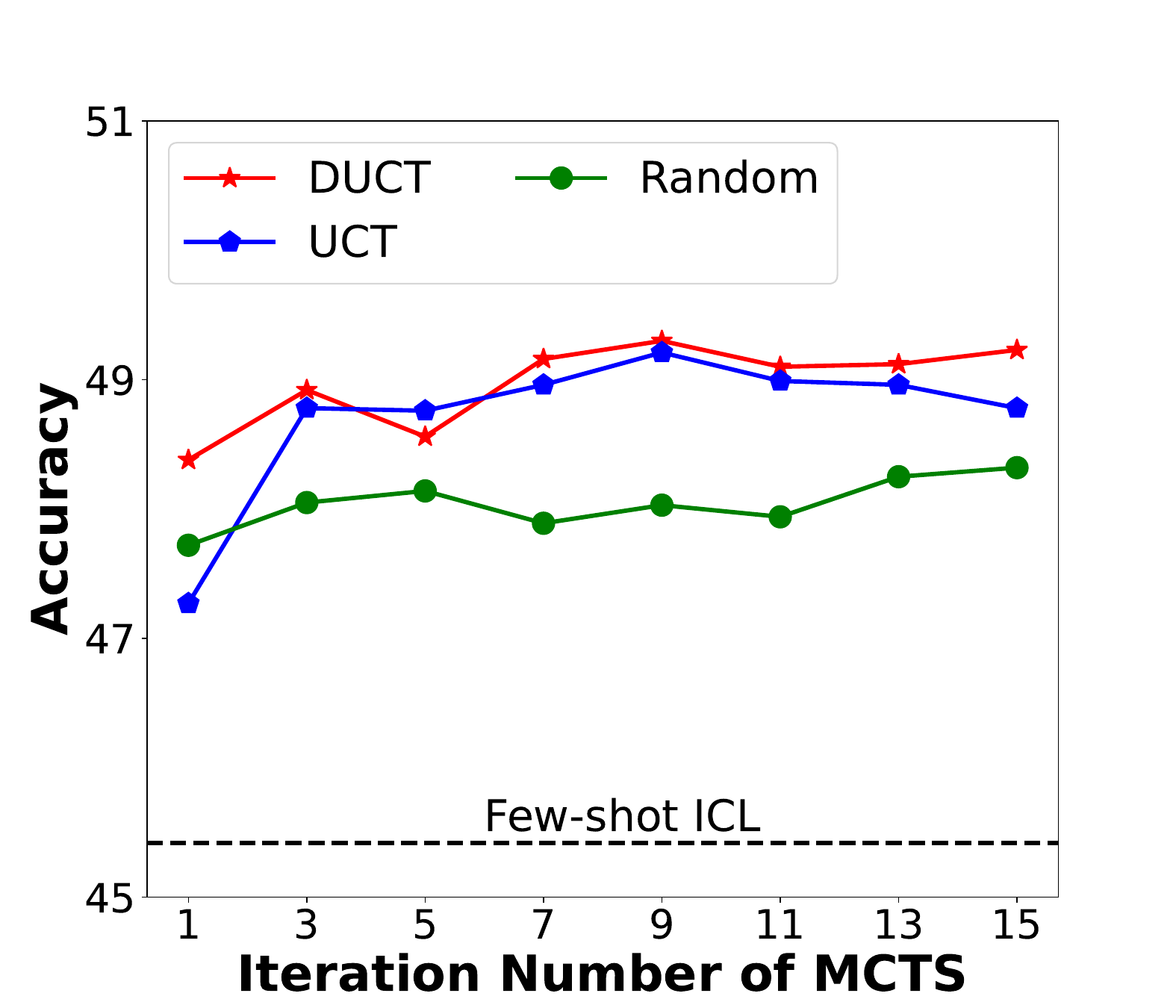}
        \caption{Llama3.1-8B}
        \label{fig:bbh-efficiency-llama}
    \end{subfigure}
    \begin{subfigure}[b]{0.49\linewidth}
        \centering
        \includegraphics[width=\textwidth]{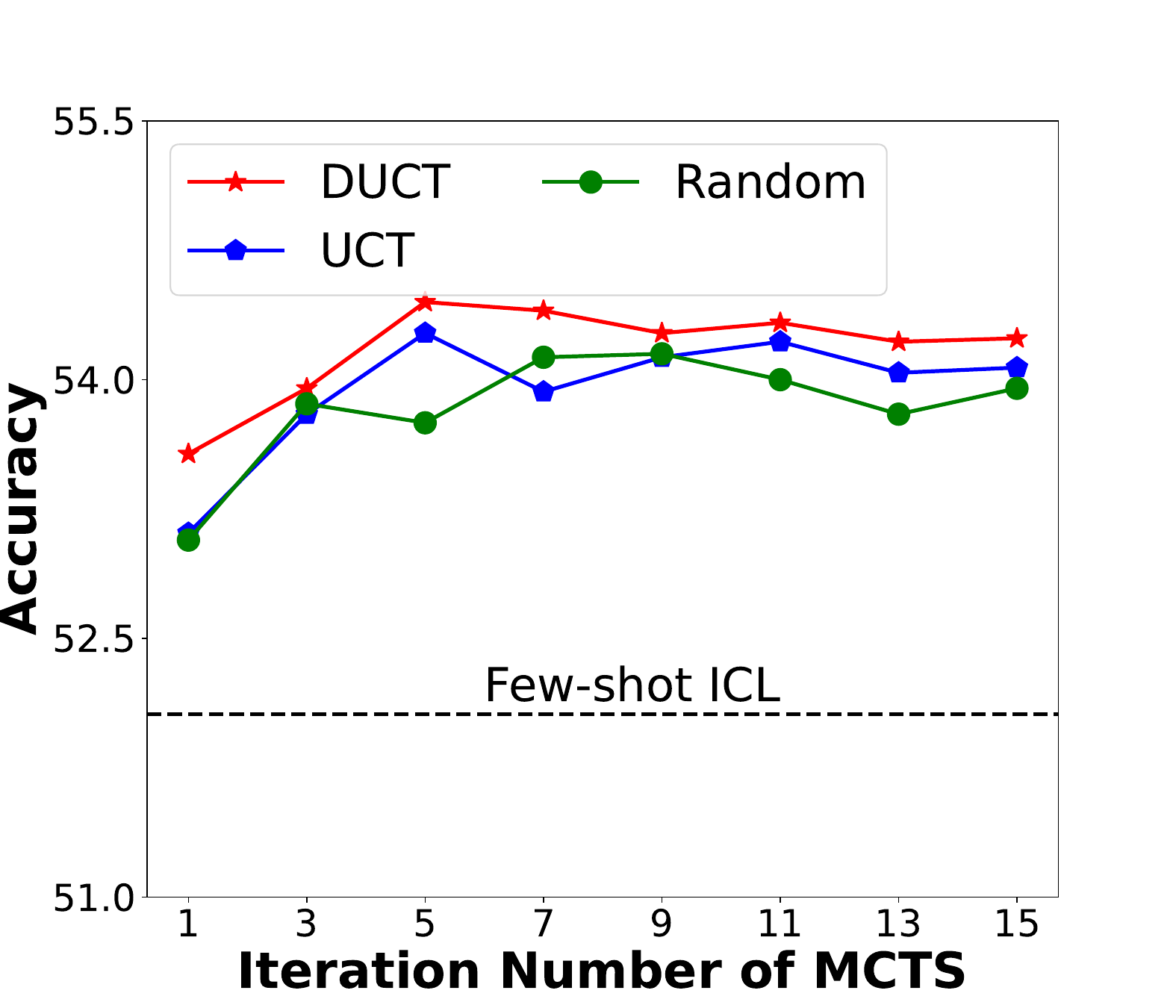}
        \caption{Qwen2.5-7B}
        \label{fig:bbh-efficiency-qwen}
    \end{subfigure}
    \caption{Accuracy on BBH with increasing numbers of iterations using the selection strategy of random, UCT, and our proposed DUCT.}
\label{fig:efficiency}
\end{figure}

In this section, we conduct an exploration efficiency analysis of our approach.
Due to limited computational resources, we run MCTS with up to 15 iterations and conduct experiments on BBH using Llama3.1-8B and Qwen2.5-7B. 
To keep the same number of inference times across different strategies, we do not use the cache method to ensure a fair comparison.
The results are presented in Figure~\ref{fig:efficiency}.
As the number of iterations increases, the performance of our approach (DUCT) quickly improves and reaches the plateau with about 9 iterations, showing its effectiveness and efficiency.
Compared with the original selection strategy UCT in MCTS, our proposed DUCT strategy achieves faster convergence and better performance.
The main reason is that it uses the demonstration information to initialize $Q$ values, achieving a more reliable estimation of the expected future reward.

\subsection{The Effect of Demonstration Selection Strategy}

\begin{figure}[t]
    \centering
    \includegraphics[width=\linewidth]{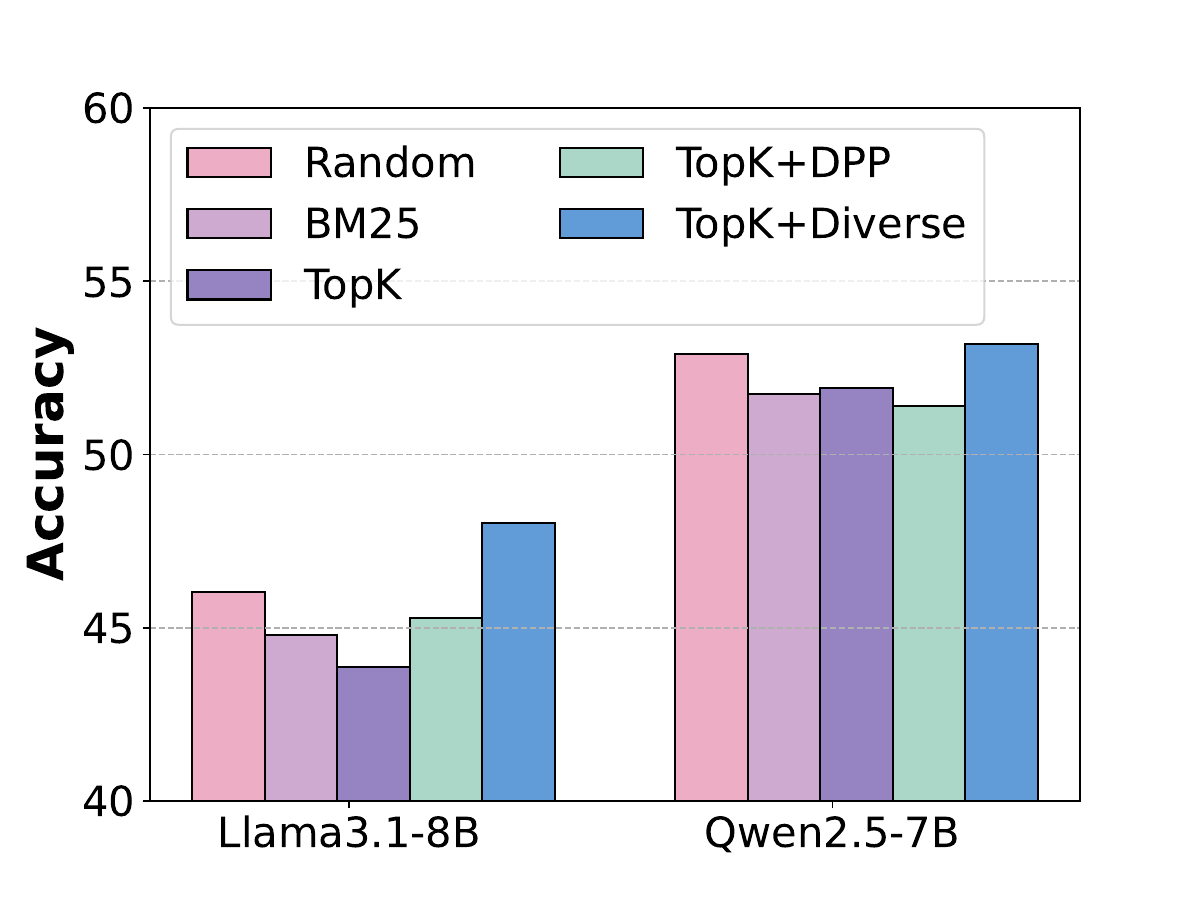}
    \caption{The Accuracy~(\%) of BBH with different demonstration selection methods.}
\label{fig:demo-select}
\end{figure}

\begin{figure}[t]
    \centering
    \includegraphics[width=\linewidth]{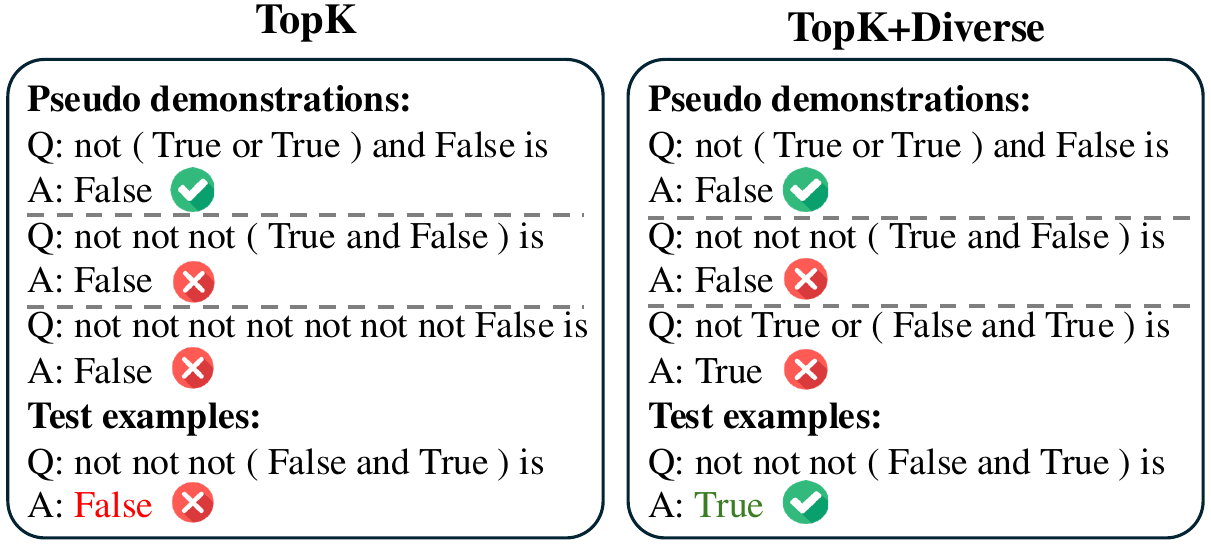}
    \caption{The error accumulation phenomenon of the similarity-based demonstration selection method.}
\label{fig:demo_select_sample}
\end{figure}

In this part, we explore the effect of various demonstration selection methods (\ie random selection, BM25~\cite{Robertson-BM25}, TopK~\cite{ACL-Liu-TopK}, and TopK+DPP~\cite{ICML-Ye-DPP}).
Specifically, the random selection method refers to randomly selecting the demonstrations for each sample.
BM25 selects demonstrations by computing the BM25 relevance score, which is a popular ranking function in information retrieval.
TopK selects demonstrations that are semantically relevant to each sample.
TopK+DPP employs a two-stage demonstration selection strategy.
In the first stage, this method retrieves $k_d$ candidates that are semantically similar to the input sample.
In the second stage, Determinantal Point Processes (DPP) are applied to simulate interactions between the query and the candidate samples to select a diverse set of demonstrations.

The experimental results are illustrated in Figure~\ref{fig:demo-select}.
It can be observed that the similarity-based demonstration selection methods~(\ie BM25 and TopK) perform worse than the random selection method.
This can be attributed to the fact that these methods tend to use samples with the same label as demonstrations, which can lead to incorrect predictions and subsequent error accumulation.
As shown in Figure~\ref{fig:demo_select_sample}, selecting semantically similar samples typically results in demonstrations that have the same label.
Due to the copying phenomenon of ICL~\cite{arXiv-Olsson-copying-ICL}, LLMs can exhibit majority bias~\cite{ICML-Zhao-calibration}, which results in generating answers that are frequent in the demonstrations. 
Furthermore, these incorrectly predicted demonstrations can mislead subsequent predictions made by the LLMs.
To address this issue, we select samples with more diverse labels, allowing LLMs to make predictions without being influenced by the labels of the demonstrations.
Experimental results indicate that our proposed TopK+Diverse demonstration selection method consistently achieves the best performance across different LLMs, confirming the effectiveness of incorporating label diversity into the demonstration selection process for ZS-ICL.
\section{Conclusion}

In this paper, we introduce \OURS, a strategic \textit{planning} approach for ZS-ICL that utilizes MCTS to search for the optimal problem-solving sequence.
To achieve effective and efficient $Q$ value estimation, we propose a novel demonstration-aware $Q$-value function that aims to enhance the \textit{selection} phase and accelerate the \textit{expansion} and \textit{simulation} phases in MCTS.
Experimental results demonstrate that our approach consistently outperforms existing ZS-ICL methods and even performs better than ICL with human-annotated demonstrations.
Overall, our work highlights the importance of planning for ZS-ICL in real-world scenarios, paving the way for more effective deployment of LLMs.
\section{Limitations}
\label{sec-limitations}

In this work, we employ the MCTS algorithm for planning the problem-solving path in ZS-ICL.
More advanced planning algorithms remain to be explored.
We estimate the expected future rewards by simulation and back-propagation.
This is quite time-consuming for ZS-ICL since each simulated state requires the LLM to perform one inference for reward calculation.
A promising direction for future work is to train a value model for efficient evaluation~\cite{arXiv-Wang-Q*}.
In addition, due to limitations in computational resources, we only conduct experiments on several representative LLMs.

\section*{Acknowledgements}
This work was partially supported by National Natural Science Foundation of China under Grant No. 92470205 and 62222215. Xin Zhao is the corresponding author.

\bibliography{newbib}

\clearpage
\appendix

\clearpage
\section{More experiments}

\subsection{The Effect of Larger Models}

\begin{table}[t]
\centering
\resizebox{\linewidth}{!}{
    \begin{tabular}{l|ccc}
    \toprule
    \textbf{Model}               & Gemma2-9B & Qwen2.5-14B & Qwen2.5-32B \\
    \midrule
    Zero-shot           & 47.58 & 56.61 & 59.74 \\
    Few-shot            & 53.87 & 59.37 & 64.80 \\
    Self-ICL            & 45.22 & 54.35 & 57.30 \\
    DAIL                & 38.89 & 57.88 & 62.13 \\
    \midrule
    \OURS               & \underline{55.89} & \underline{61.37} & \underline{66.72} \\
    \OURS w/o cache     & \textbf{57.01} & \textbf{63.35} & \textbf{68.44} \\
\bottomrule
\end{tabular}
}
\caption{Performance comparison across larger open-source LLMs on BBH. The best method in each group is marked in \textbf{bold}, and the second-best method is marked with an \underline{underline}.}
\label{tab:larger_models}
\end{table}

In this part, we conduct experiments on larger open-source LLMs (\ie Gemma2-9B, Qwen2.5-14B, and Qwen2.5-32B) on BBH.
As illustrated in Table~\ref{tab:larger_models}, \OURS can outperform other competitive ZS-ICL baselines and even surpass few-shot prompting with human-labeled demonstrations on larger LLMs.
This further demonstrates the effectiveness of our proposed method.

\subsection{Experiments on Generation Tasks}

\begin{table}[t]
\centering
\tiny
\resizebox{\linewidth}{!}{
    \begin{tabular}{l|cc}
    \toprule
    \textbf{Method}    & MultiArith & Last Letter \\
    \midrule
    Zero-shot          & 94.44 & 86.00 \\
    Few-shot           & 94.44 & 88.00 \\
    Self-ICL           & 96.67 & 86.67 \\
    DAIL               & 96.11 & 89.33 \\
    \midrule
    \OURS              & \underline{97.22} & \underline{91.33} \\
    \OURS w/o cache    & \textbf{98.33} & \textbf{92.00} \\
\bottomrule
\end{tabular}
}
\caption{Performance comparison on generation tasks using GPT-4o-mini. The best method in each group is marked in \textbf{bold}, and the second-best method is marked with an \underline{underline}.}
\label{tab:generation_tasks}
\end{table}

To confirm the effectiveness of our proposed method on generation tasks, we conduct experiments on two types of generation tasks (\ie math reasoning and symbolic reasoning) using GPT-4o-mini.
Specifically, for the math reasoning problem, we use Multiarith~\cite{EMNLP-Multiarith-dataset} for evaluation.
For the symbolic reasoning problem, we use Last Letter Concatenation~\cite{NIPS-Last-letter-concat} for evaluation.
As shown in Table~\ref{tab:generation_tasks}, \OURS consistently outperforms other methods on these generation tasks.
This further highlights the importance of strategically planning problem-solving trajectories and demonstrates the effectiveness of \OURS.

\subsection{Hyper-parameters Analysis}
\label{app:hyper}

\begin{figure}[t]
    \centering
    \begin{subfigure}[b]{0.49\linewidth}
        \centering
        \includegraphics[width=\textwidth]{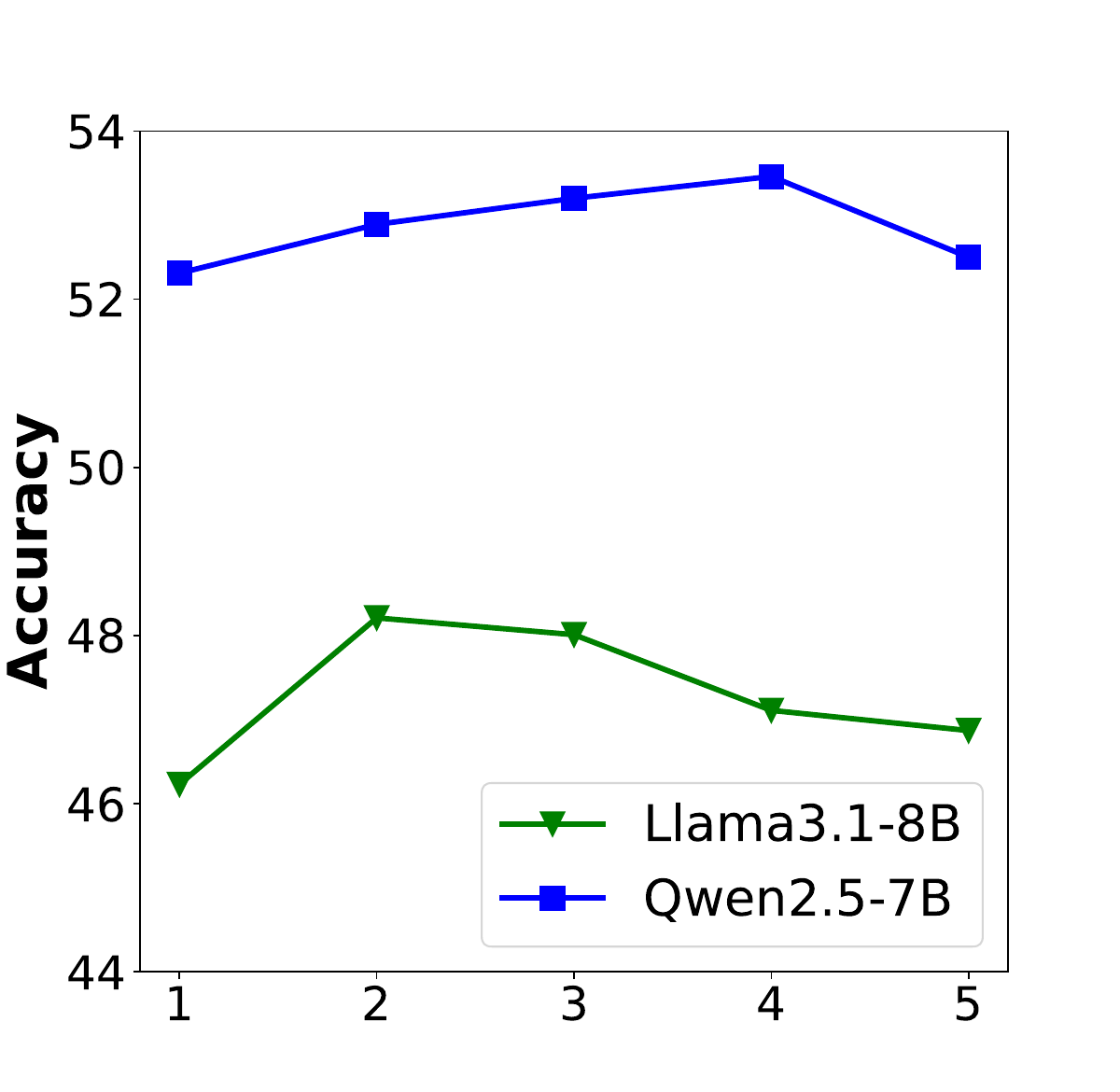}
        \caption{The Number of Node Expansion on BBH.}
        \label{fig:bbh-expansion_size}
    \end{subfigure}
    \begin{subfigure}[b]{0.49\linewidth}
        \centering
        \includegraphics[width=\textwidth]{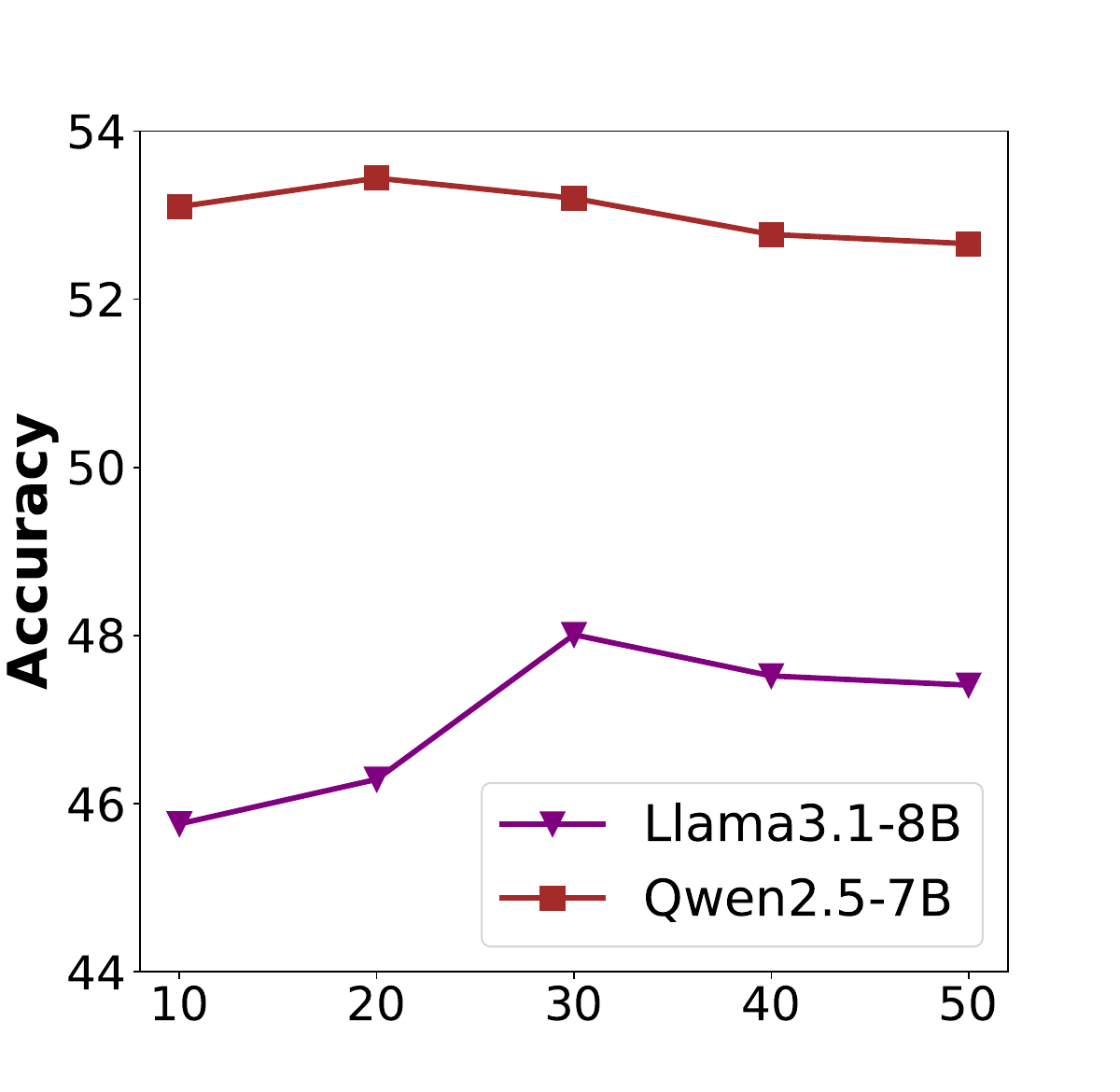}
        \caption{The Number of the Retrieved Candidates on BBH.}
        \label{fig:bbh-retrieve_candidate}
    \end{subfigure}
    \vskip\baselineskip
    \begin{subfigure}[b]{0.49\linewidth}
        \centering
        \includegraphics[width=\textwidth]{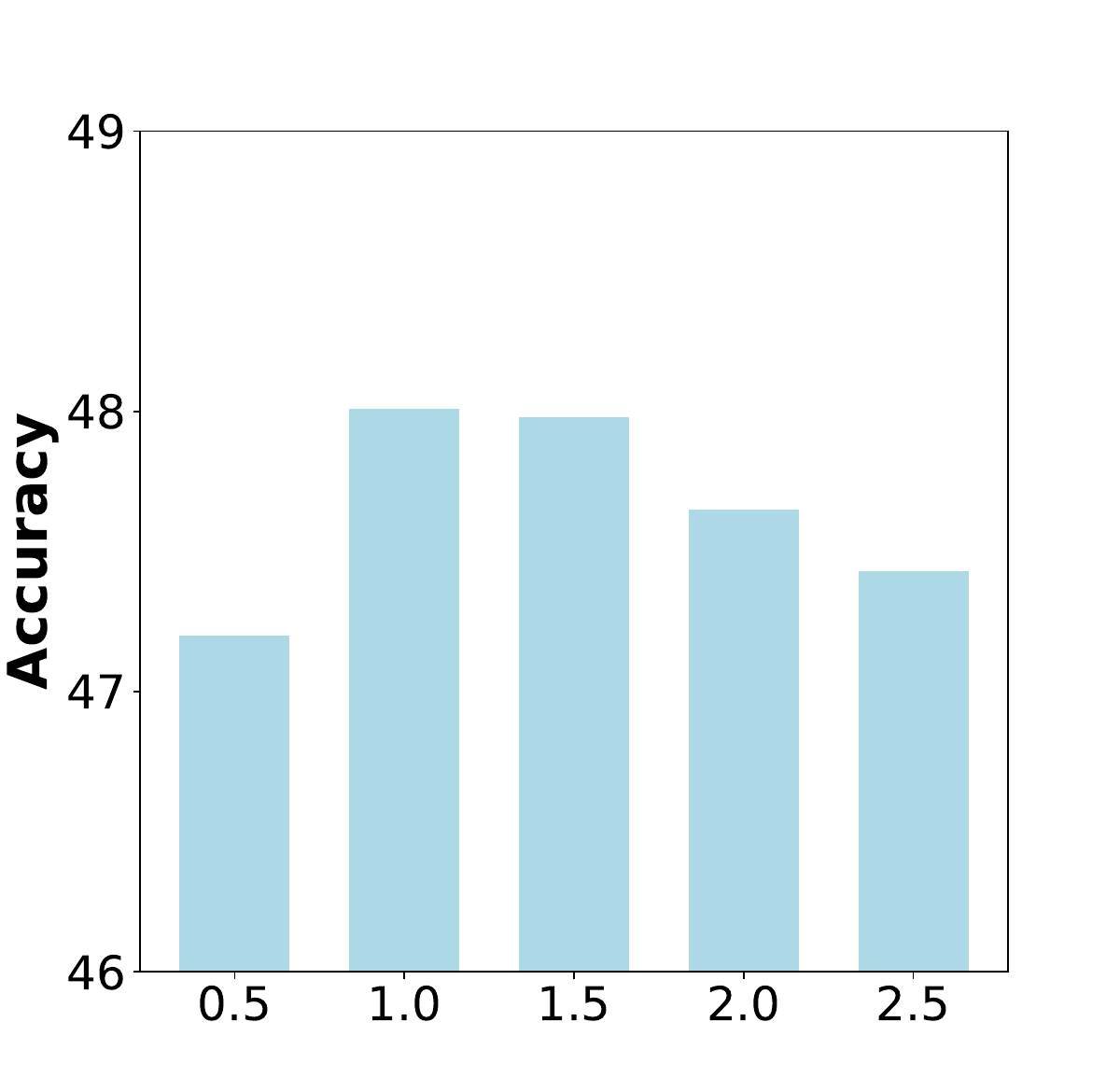}
        \caption{Analysis of the Hyper-parameter $w_a$ on BBH using Llama3.1-8B.}
        \label{fig:bbh-wa}
    \end{subfigure}
    \begin{subfigure}[b]{0.49\linewidth}
        \centering
        \includegraphics[width=\textwidth]{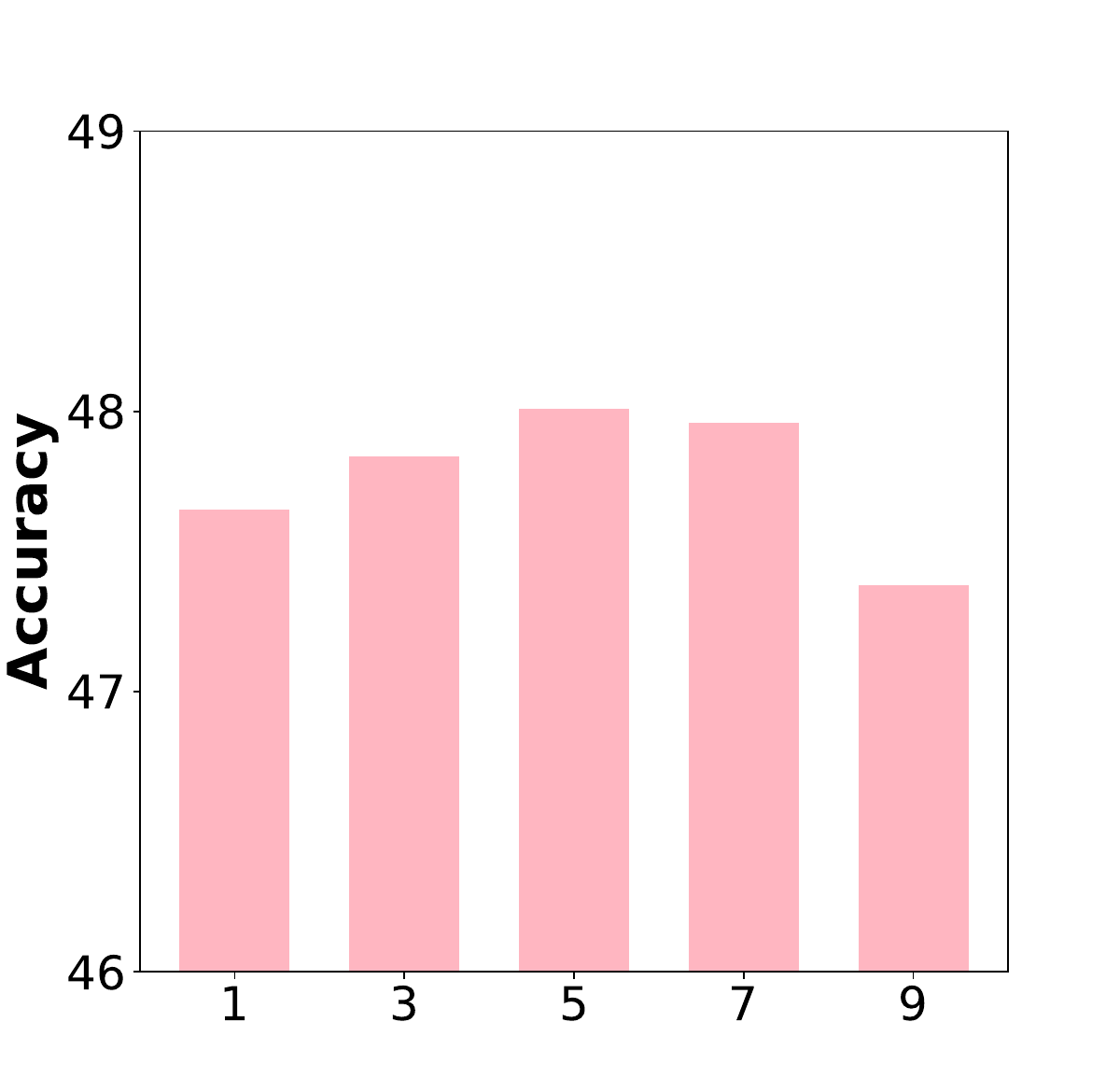}
        \caption{Analysis of the Hyper-parameter $w_Q$ on BBH using Llama3.1-8B.}
        \label{fig:bbh-wq}
    \end{subfigure}
\caption{Performance comparison \emph{w.r.t.} the number of expansion nodes, retrieved candidates, balance constant between exploration and exploitation, and balance constant of the initial and updated $Q$-value of \OURS.}
\label{fig:hyper-para}
\end{figure}

\OURS includes a few hyper-parameters to tune.
In this section, we report the tuning results of four hyper-parameters on BBH: the expansion size~(\ie $k_a$), the number of retrieved candidates~(\ie $k_d$), the balance constant between exploration and exploitation~(\ie $w_a$), and the balance constant between the initial and updated $Q$ value~(\ie $w_Q$).
The results are presented in Figure~\ref{fig:hyper-para}.
We observe that \OURS achieves the best performance when the expansion size is set to 3.
If the expansion size is too large, it may introduce suboptimal states which can lead to a shift in the exploration direction.
Conversely, if the expansion size is too small, the tree will not expand sufficiently due to the absence of some important nodes.
On the other hand, we find that the number of retrieved candidates cannot be too small or too large.
If the number is too small, some important examples can be overlooked, which results in performance degradation.
In contrast, if the number is too large, irrelevant examples can be introduced during the calculation of $Q_0$, leading to a deviation in the exploration direction.
For $w_a$ and $w_Q$, we find that they have minimal effect on performance, with the best results achieved when $w_a$ is set to 5 and $w_Q$ is set to 1.

\section{Details of Experimental Cost}
\label{sec-detailed-experiments-cost}

\begin{table}[t]
\centering
\resizebox{\linewidth}{!}{
\begin{tabular}{l|c|c|c}
\toprule
\textbf{Method} & \begin{tabular}[c]{@{}c@{}}\textbf{Input} \\ \textbf{Tokens}\end{tabular} & \begin{tabular}[c]{@{}c@{}}\textbf{Output} \\ \textbf{Tokens}\end{tabular} & \textbf{Cost} \\
\midrule
\midrule
\multicolumn{4}{c}{\textit{BBH}} \\
\midrule
\midrule
Zero-shot         & 764K      & 15K & 0.12 \\
Few-shot          & 2324K     & 15K & 0.36 \\
Self-ICL          & 3537K     & 1197K & 1.25 \\
DAIL              & 2567K     & 15K & 0.39 \\
\OURS             & 3418K     & 20K & 0.52 \\
\OURS w/o cache   & 37836K    & 220K & 5.81 \\
\midrule
\midrule
\multicolumn{4}{c}{\textit{MMLU}} \\
\midrule
\midrule
Zero-shot         & 1491K & 14K & 0.23 \\
Few-shot          & 7761K & 14K & 1.17 \\
Self-ICL          & 12364K & 3435K & 3.92 \\
DAIL              & 7728K  & 14K & 1.17 \\
\OURS             & 28881K & 52K & 4.36 \\
\OURS w/o cache   & 114446K & 207K & 17.29 \\
\bottomrule
\end{tabular}
}
\caption{The number of consumed tokens and the cost (in US dollars) on BBH and MMLU using GPT-4o-mini.}
\label{tab:cost}
\end{table}

In this section, we present detailed information on the experimental costs of GPT-4o-mini in Table~\ref{tab:cost}.
\section{Detailed Results}
\label{sec-detailed_results}

In this part, we report the detailed experimental results on BBH and MMLU across four LLMs in Table~\ref{tab:bbh-all1}, \ref{tab:bbh-all2}, \ref{tab:mmlu-all1}, and \ref{tab:mmlu-all2}.
\section{Example Prompts}
\label{sec-prompt}

In this part, we present the example prompt for BBH and MMLU.

\begin{center}
\begin{tcolorbox}[colback=blue!5!white,colframe=blue!55!black,width=0.98\linewidth,title={\textit{Example Prompt for BBH}}]
{
    {
    \small
    \textbf{Pseudo demonstrations:} \\
    Q: Which statement is sarcastic? \\
    Options:\\
    (A) But his eyes were on the ball, shouldn't be a red \\
    (B) But his cleats were on the ball, shouldn't be a red \\
    A: (B) \\ 
    Q: Is the following sentence plausible? ``John Carlson scored in the third period.'' \\
    A: yes \\
    Q: Is the following sentence plausible? ``Elias Lindholm beat the buzzer.'' \\
    A: no \\
    }
    {
    \small
    \textbf{Test example:} \\
    Q: Is the following sentence plausible? ``Marcelo got on the end of a through ball.'' \\
    A: \textcolor{red}{yes}
    }
}
\end{tcolorbox}
\end{center}

\begin{center}
\begin{tcolorbox}[colback=blue!5!white,colframe=blue!55!black,width=0.98\linewidth,title={\textit{Example Prompt for MMLU}}]
{
    {
    \small
    \textbf{Pseudo demonstrations:} \\
    Question: Objects that absorb light appear A.black    B.white    C.dark    D.bright \\
    Answer: A \\
    Question: Of the following, most children will develop which skill first? \\
    A.write with a pencil    B.cut with a knife    C.say a sentence    D.clap their hands \\
    Answer: D \\
    Question: Which of the following most accurately explains why a pool with water temperature of 82 degrees may feel cool to a person who has been sunbathing, yet warm to a person who has been inside in the air conditioning? \\
    A.Sensory restriction    B.Perceptual constancy    C.Relative clarity    D.Sensory adaptation \\
    Answer: B \\
    Question: What part of the human body does a gastroenterologist examine? \\
    A.Brain    B.Skeleton    C.Stomach    D.Nose \\
    Answer: A \\
    }
    {
    \small
    \textbf{Test example:} \\
    Question: Paper will burn at approximately what temperature in Fahrenheit? \\
    A.986 degrees    B.2125 degrees    C.3985 degrees    D.451 degrees \\
    Answer: \textcolor{red}{D}
    }
}
\end{tcolorbox}
\end{center}
\clearpage
\algnewcommand{\Inputs}[1]{%
  \State \textbf{Inputs:}
  \Statex \hspace*{\algorithmicindent}\parbox[t]{0.98\linewidth}{\raggedright #1}
}
\algnewcommand{\Initialize}[1]{%
  \State \textbf{Initialize:}
  \Statex \hspace*{\algorithmicindent}\parbox[t]{0.98\linewidth}{\raggedright #1}
}

\begin{algorithm*}[t]
\centering
\caption{$\operatorname{\OURS}(\mathcal X, s_0, t_\theta, r_\theta, k_a, \tau, \epsilon)$}
\label{alg:mcts}
\begin{minipage}{1\linewidth} 
\small
\begin{algorithmic}
    \Inputs{
    Problem set $\mathcal X = \{x_i\}_{i=1}^n $, Initial state $s_0$, deterministic state transition function $t_\theta$,  \\
     reward function $r_\theta$, action expansion number $k_a$, iteration number $\tau$, action cache threshold $\epsilon$
    }
    \Initialize{
    State to action mapping $A : \mathcal S \mapsto \mathcal A$, children mapping $\text{ch} : \mathcal S \times \mathcal A \mapsto \mathcal S$, rewards $r : \mathcal S \times \mathcal A \mapsto \mathbb R$, \\
    visited counter $\mathcal{N} : \mathcal S \mapsto \mathbb N$, action cache $\text{ac} : \varnothing$, State-action value function $Q : \mathcal S \times \mathcal A \mapsto \mathbb R$, \\
    Demonstration-aware state-action value function $\text{DQ} : \mathcal S \times \mathcal A \mapsto \mathbb R$
    }
    \For {$n \gets 0, \dots, \tau - 1$}
        \State $t \gets 0$
        \While {$\mathcal{N}(s_t)$ > 0} \Comment{selection}
            \State Select $a_t$ from $A(s_t)$ with the highest DUCT score based on Equation~\ref{eq:selection}
            \State $s_{t + 1} \gets \text{ch}(s_t, a_t)$, $\mathcal{N}(s_{t}) \gets \mathcal{N}(s_{t}) + 1$
            \State $t \gets t+1$
        \EndWhile
        \While {$s_t$ is not a terminal state} \Comment{expansion and simulation}
            \State Select $a_t$ from $A(s_t)$ with the top-$k_a$ $\text{DQ}$ score based on Equation~\ref{eq:Q-func}
            \For {$i \gets 1, \dots, k_a$}
                \If {$a_t^i$ in $\text{Ac}$} 
                    \State Read $(x_i, \hat{y_i})$ from $\text{Ac}$                
                \Else
                    \State Perform zero-shot in-context learning for $x_i$ to generate $\hat{y_i}$
                    \State $r_{t}^i \gets r_\theta(s_{t}^i, a_t^i)$         
                    \If {$\text{DQ}_t^i > \epsilon$} 
                        \State Write $(a_i, (x_i, \hat{y_i}))$ to $\text{Ac}$
                    \EndIf
                \EndIf
                \State $s_{t+1}^i \gets t_\theta(s_t, a_t^i)$
            \EndFor
            \State $a_{t}^* \gets \mathop{\arg\max}_{a_t^i \in A(s_t)} r_{t}^i(s_{t}^i, a_t^i)$
            \State $s_{t+1} \gets \text{ch}(s_t, a_t^*)$, $\mathcal{N}(s_{t}) \gets \mathcal{N}(s_{t}) + 1$
            \State $t \gets t + 1$
            
        \EndWhile
        \State $T \gets$ the actual number of simulation steps
        \For {$t \gets T - 1, \dots, 0$} \Comment{back-propagation}
            \State Update $Q(s_t, a_t)$ with $\{r_t, r_{t+1}, \dots, r_T\}$.
        \EndFor
    \EndFor
\end{algorithmic}
\end{minipage}
\end{algorithm*}
\begin{table*}[ht]
\centering
\resizebox{\textwidth}{!}{
\begin{tabular}{l|cccccc|cccccc}
\toprule
\multirow{3.5}{*}{\textbf{BBH Tasks}} & \multicolumn{6}{c|}{\textbf{Llama3.1-8B}} & \multicolumn{6}{c}{\textbf{Qwen2.5-7B}} \\
\cmidrule(lr){2-7} \cmidrule(lr){8-13}
& ZS & FS & Self-ICL & DAIL & \OURS & \begin{tabular}[c]{@{}c@{}}\OURS\\ w/o cache\end{tabular} & ZS & FS & Self-ICL & DAIL & \OURS & \begin{tabular}[c]{@{}c@{}}\OURS\\ w/o cache\end{tabular} \\
\midrule
Boolean Expressions & 71.60 & 80.40 & 62.40 & 75.20 & 80.40 & 83.20 & 85.20 & 87.60 & 78.00 & 84.80 & 90.00 & 89.20 \\
Causal Judgement & 51.87 & 52.94 & 51.87 & 51.87 & 51.87 & 54.55 & 52.41 & 51.87 & 52.41 & 53.48 & 59.89 & 61.50 \\
Date Understanding & 50.00 & 51.20 & 36.80 & 59.20 & 53.20 & 54.40 & 61.20 & 56.80 & 54.40 & 59.20 & 58.40 & 61.60 \\
Disambiguation QA & 40.00 & 57.20 & 46.80 & 39.20 & 66.40 & 54.80 & 59.60 & 65.60 & 64.80 & 62.00 & 66.80 & 69.20 \\
Formal Fallacies & 53.20 & 53.20 & 54.40 & 52.80 & 52.00 & 54.00 & 56.80 & 57.20 & 49.20 & 58.40 & 59.60 & 57.60 \\
Geometric Shapes & 9.20 & 40.00 & 18.00 & 28.40 & 43.60 & 42.40 & 25.20 & 53.60 & 31.20 & 30.80 & 38.80 & 41.60 \\
Hyperbaton & 75.20 & 60.80 & 57.60 & 67.20 & 78.80 & 80.40 & 68.40 & 63.60 & 79.20 & 71.20 & 82.40 & 82.80 \\
Logical Deduction(five objects) & 38.00 & 37.60 & 29.20 & 37.60 & 38.80 & 40.00 & 50.00 & 50.00 & 47.60 & 45.60 & 50.00 & 50.80 \\
Logical Deduction(seven objects) & 37.60 & 30.40 & 21.60 & 42.00 & 47.20 & 48.00 & 48.40 & 51.20 & 43.60 & 49.20 & 50.80 & 52.80 \\
Logical Deduction(three objects) & 52.40 & 50.80 & 36.80 & 48.40 & 53.60 & 56.40 & 75.20 & 76.40 & 74.40 & 73.60 & 74.00 & 77.20 \\
Movie Recommendation & 77.60 & 84.40 & 47.20 & 78.80 & 76.00 & 83.60 & 74.40 & 74.80 & 63.60 & 74.40 & 76.00 & 77.20 \\
Navigate & 42.00 & 42.00 & 42.00 & 42.00 & 53.20 & 53.20 & 48.00 & 42.00 & 42.00 & 42.40 & 61.20 & 65.20 \\
Penguins in a Table & 40.41 & 43.15 & 32.19 & 39.04 & 50.68 & 46.58 & 57.53 & 58.22 & 52.74 & 54.79 & 56.16 & 54.79 \\
Reasoning about Colored Objects & 45.60 & 44.00 & 12.80 & 40.00 & 50.00 & 48.40 & 56.00 & 56.80 & 34.40 & 54.40 & 53.20 & 57.20 \\
Ruin Names & 40.00 & 58.80 & 40.40 & 44.40 & 42.40 & 44.00 & 46.40 & 57.20 & 50.80 & 50.80 & 53.20 & 52.80 \\
Salient Translation Error Detection & 20.40 & 35.20 & 20.80 & 30.00 & 36.00 & 36.40 & 34.40 & 46.80 & 36.40 & 45.20 & 48.00 & 49.60 \\
Snarks & 51.69 & 57.30 & 48.88 & 53.93 & 58.43 & 61.80 & 66.85 & 77.53 & 73.03 & 74.16 & 76.97 & 75.84 \\
Sports Understanding & 46.00 & 46.00 & 46.00 & 46.00 & 67.60 & 70.80 & 65.20 & 46.00 & 46.00 & 46.00 & 62.40 & 67.20 \\
Temporal Sequences & 33.60 & 11.20 & 34.40 & 5.60 & 2.80 & 2.00 & 17.60 & 21.60 & 22.80 & 13.20 & 18.80 & 16.40 \\
Tracking Shuffled Objects(five objs) & 16.00 & 15.60 & 14.00 & 13.60 & 12.40 & 12.40 & 16.80 & 19.20 & 19.20 & 14.00 & 10.80 & 12.00 \\
Tracking Shuffled Objects(seven objs) & 10.80 & 12.80 & 15.60 & 9.20 & 10.00 & 8.80 & 14.80 & 17.20 & 16.80 & 13.20 & 12.80 & 10.40 \\
Tracking Shuffled Objects(three objs) & 25.60 & 35.20 & 30.00 & 32.00 & 29.60 & 31.60 & 29.20 & 27.20 & 24.80 & 28.40 & 20.40 & 20.00 \\
Web of Lies & 48.80 & 48.80 & 48.80 & 48.80 & 54.40 & 53.60 & 48.80 & 48.80 & 48.80 & 48.80 & 52.80 & 53.60 \\
\midrule
\textbf{All Tasks~(avg)} & 42.32 & 45.42 & 36.65 & 42.66 & 48.01 & \textbf{48.56} & 49.99 & 52.06 & 47.63 & 49.46 & 53.20 & \textbf{54.27} \\
\bottomrule
\end{tabular}
}
\caption{Detailed Results on BBH for Llama3.1-8B and Qwen2.5-7B. ``ZS'' and ``FS'' denote zero-shot and few-shot prompting, respectively.}
\label{tab:bbh-all1}
\end{table*}
\begin{table*}[ht]
\centering
\resizebox{\textwidth}{!}{
\begin{tabular}{l|cccccc|cccccc}
\toprule
\multirow{3.5}{*}{\textbf{BBH Tasks}} & \multicolumn{6}{c|}{\textbf{Mistral-7B}} & \multicolumn{6}{c}{\textbf{GPT-4o-mini}} \\
\cmidrule(lr){2-7} \cmidrule(lr){8-13}
& ZS & FS & Self-ICL & DAIL & \OURS & \begin{tabular}[c]{@{}c@{}}\OURS\\ w/o cache\end{tabular} & ZS & FS & Self-ICL & DAIL & \OURS & \begin{tabular}[c]{@{}c@{}}\OURS\\ w/o cache\end{tabular} \\
\midrule
Boolean Expressions & 76.00 & 82.40 & 73.20 & 74.40 & 79.20 & 84.80 & 51.20 & 63.20 & 59.20 & 57.20 & 49.20 & 57.20 \\
Causal Judgement & 50.80 & 55.08 & 52.94 & 53.48 & 53.48 & 52.41 & 59.89 & 66.31 & 62.57 & 60.96 & 64.17 & 64.71 \\
Date Understanding & 50.00 & 52.00 & 44.80 & 54.00 & 57.60 & 60.80 & 22.00 & 56.00 & 48.40 & 62.40 & 61.60 & 61.20 \\
Disambiguation QA & 49.20 & 60.80 & 57.60 & 54.40 & 62.00 & 68.40 & 37.60 & 50.00 & 51.20 & 50.80 & 52.00 & 53.20 \\
Formal Fallacies & 53.60 & 47.20 & 48.00 & 50.00 & 48.80 & 52.40 & 52.80 & 63.20 & 58.00 & 60.80 & 60.00 & 60.80 \\
Geometric Shapes & 9.20 & 38.40 & 10.80 & 27.60 & 41.20 & 40.00 & 31.20 & 34.00 & 30.80 & 38.00 & 40.80 & 40.00 \\
Hyperbaton & 74.80 & 76.40 & 70.00 & 54.40 & 74.80 & 76.40 & 51.60 & 61.60 & 68.80 & 69.60 & 86.00 & 88.00 \\
Logical Deduction(five objects) & 27.60 & 34.40 & 27.60 & 31.20 & 36.00 & 37.20 & 18.40 & 24.80 & 22.40 & 29.60 & 34.00 & 32.40 \\
Logical Deduction(seven objects) & 20.40 & 26.00 & 21.60 & 28.00 & 36.00 & 34.80 & 15.20 & 17.20 & 18.00 & 17.20 & 19.60 & 20.00 \\
Logical Deduction(three objects) & 43.20 & 46.00 & 47.60 & 42.80 & 52.80 & 52.80 & 35.20 & 76.00 & 69.20 & 67.20 & 75.60 & 74.40 \\
Movie Recommendation & 39.20 & 67.60 & 46.40 & 66.40 & 78.80 & 80.80 & 31.20 & 70.40 & 56.00 & 69.20 & 77.20 & 77.20 \\
Navigate & 45.60 & 50.40 & 52.00 & 48.80 & 55.60 & 54.00 & 50.80 & 68.80 & 69.60 & 68.40 & 65.60 & 67.60 \\
Penguins in a Table & 32.88 & 34.25 & 32.88 & 41.10 & 38.36 & 40.41 & 48.63 & 58.22 & 54.79 & 58.22 & 53.42 & 54.79 \\
Reasoning about Colored Objects & 35.20 & 31.60 & 27.20 & 28.00 & 28.00 & 27.60 & 62.80 & 68.00 & 69.60 & 67.20 & 71.60 & 72.80 \\
Ruin Names & 40.00 & 43.60 & 35.20 & 31.20 & 54.80 & 53.60 & 81.60 & 85.20 & 84.80 & 83.20 & 87.20 & 87.20 \\
Salient Translation Error Detection & 14.80 & 32.40 & 21.20 & 32.80 & 34.80 & 30.40 & 59.20 & 58.40 & 55.20 & 56.00 & 58.40 & 59.60 \\
Snarks & 50.00 & 55.62 & 43.26 & 47.75 & 53.93 & 52.81 & 64.04 & 59.55 & 53.93 & 67.42 & 71.91 & 70.79 \\
Sports Understanding & 47.20 & 72.80 & 60.40 & 66.00 & 72.40 & 75.20 & 51.60 & 84.00 & 78.00 & 82.00 & 86.40 & 87.20 \\
Temporal Sequences & 21.20 & 18.80 & 19.20 & 14.40 & 8.40 & 8.40 & 27.60 & 77.60 & 74.00 & 71.20 & 83.20 & 83.60 \\
Tracking Shuffled Objects(five objs) & 19.20 & 19.20 & 22.00 & 23.20 & 16.00 & 12.00 & 16.00 & 14.00 & 17.20 & 12.80 & 14.40 & 14.00 \\
Tracking Shuffled Objects(seven objs) & 13.20 & 13.20 & 11.60 & 11.60 & 8.00 & 9.60 & 10.80 & 9.20 & 11.60 & 11.20 & 8.00 & 7.60 \\
Tracking Shuffled Objects(three objs) & 33.20 & 32.40 & 33.20 & 35.60 & 31.20 & 34.40 & 35.60 & 28.80 & 33.20 & 27.20 & 29.60 & 30.40 \\
Web of Lies & 48.80 & 52.40 & 51.20 & 56.40 & 55.20 & 51.60 & 49.20 & 50.40 & 53.20 & 56.40 & 52.80 & 52.00 \\
\midrule
\textbf{All Tasks (avg)}                & 38.76 & 45.31 & 39.48 & 42.15 & 46.83 & \textbf{47.43} & 41.30 & 53.84 & 51.97 & 53.77 & 56.41 & \textbf{57.03} \\
\bottomrule
\end{tabular}
}
\caption{Detailed results on BBH tasks for Mistral-7B and GPT-4o-mini. ``ZS'' and ``FS'' denote zero-shot and few-shot prompting, respectively.}
\label{tab:bbh-all2}
\end{table*}
\begin{table*}[ht]
\centering
\resizebox{\textwidth}{!}{
\begin{tabular}{l|cccccc|cccccc}
\toprule
\multirow{3.5}{*}{\textbf{MMLU Tasks}} & \multicolumn{6}{c|}{\textbf{Llama3.1-8B}} & \multicolumn{6}{c}{\textbf{Qwen2.5-7B}} \\
\cmidrule(lr){2-7} \cmidrule(lr){8-13}
& ZS & FS & Self-ICL & DAIL & \OURS & \begin{tabular}[c]{@{}c@{}}\OURS\\ w/o cache\end{tabular} & ZS & FS & Self-ICL & DAIL & \OURS & \begin{tabular}[c]{@{}c@{}}\OURS\\ w/o cache\end{tabular} \\
\midrule
abstract algebra & 30.00 & 29.00 & 25.00 & 34.00 & 32.00 & 34.00 & 45.00 & 50.00 & 42.00 & 44.00 & 53.00 & 50.00 \\
anatomy & 57.78 & 58.52 & 53.33 & 60.74 & 60.00 & 62.96 & 70.37 & 70.37 & 68.15 & 68.89 & 68.89 & 68.89 \\
astronomy & 62.50 & 63.82 & 55.26 & 62.50 & 68.42 & 67.11 & 80.26 & 83.55 & 77.63 & 82.24 & 84.87 & 84.87 \\
business ethics & 55.00 & 67.00 & 58.00 & 66.00 & 72.00 & 68.00 & 71.00 & 74.00 & 66.00 & 75.00 & 72.00 & 76.00 \\
clinical knowledge & 64.53 & 69.81 & 57.74 & 70.19 & 72.08 & 73.96 & 79.25 & 78.11 & 74.72 & 79.25 & 78.49 & 80.38 \\
college biology & 70.14 & 77.78 & 59.72 & 74.31 & 78.47 & 79.86 & 88.19 & 84.72 & 79.86 & 86.11 & 84.72 & 85.42 \\
college chemistry & 41.00 & 44.00 & 37.00 & 48.00 & 45.00 & 46.00 & 51.00 & 55.00 & 53.00 & 53.00 & 54.00 & 48.00 \\
college computer science & 47.00 & 53.00 & 42.00 & 53.00 & 49.00 & 53.00 & 61.00 & 67.00 & 58.00 & 66.00 & 64.00 & 62.00 \\
college mathematics & 38.00 & 39.00 & 26.00 & 32.00 & 39.00 & 36.00 & 38.00 & 47.00 & 44.00 & 49.00 & 49.00 & 43.00 \\
college medicine & 57.23 & 61.85 & 49.71 & 65.32 & 63.58 & 64.74 & 69.94 & 66.47 & 65.90 & 68.79 & 69.36 & 70.52 \\
college physics & 42.16 & 38.24 & 37.25 & 49.02 & 46.08 & 41.18 & 45.10 & 50.98 & 41.18 & 49.02 & 50.00 & 47.06 \\
computer security & 66.00 & 77.00 & 64.00 & 77.00 & 70.00 & 76.00 & 81.00 & 84.00 & 78.00 & 83.00 & 80.00 & 79.00 \\
conceptual physics & 51.91 & 53.62 & 44.68 & 53.62 & 55.32 & 56.17 & 71.06 & 71.06 & 66.81 & 70.64 & 70.21 & 69.79 \\
econometrics & 34.21 & 43.86 & 34.21 & 45.61 & 47.37 & 45.61 & 55.26 & 57.89 & 55.26 & 63.16 & 62.28 & 65.79 \\
electrical engineering & 44.83 & 58.62 & 50.34 & 61.38 & 66.21 & 65.52 & 67.59 & 73.10 & 62.76 & 71.72 & 71.72 & 71.72 \\
elementary mathematics & 34.66 & 37.30 & 30.42 & 36.77 & 40.21 & 41.80 & 56.08 & 67.72 & 62.96 & 70.11 & 68.52 & 70.90 \\
formal logic & 38.89 & 47.62 & 34.92 & 48.41 & 44.44 & 44.44 & 49.21 & 57.14 & 50.00 & 58.73 & 60.32 & 60.32 \\
global facts & 24.00 & 29.00 & 21.00 & 27.00 & 34.00 & 35.00 & 51.00 & 40.00 & 45.00 & 45.00 & 44.00 & 47.00 \\
high school biology & 76.77 & 77.74 & 69.35 & 82.26 & 78.39 & 81.61 & 84.52 & 87.42 & 83.23 & 87.10 & 90.00 & 88.39 \\
high school chemistry & 46.80 & 48.77 & 37.93 & 49.26 & 52.71 & 53.69 & 67.98 & 67.49 & 61.58 & 69.46 & 65.02 & 69.95 \\
high school computer science & 56.00 & 63.00 & 54.00 & 62.00 & 62.00 & 63.00 & 80.00 & 81.00 & 80.00 & 82.00 & 84.00 & 83.00 \\
high school european history & 70.91 & 75.15 & 70.91 & 73.94 & 75.76 & 78.18 & 81.82 & 81.82 & 80.61 & 79.39 & 80.61 & 81.82 \\
high school geography & 76.26 & 78.28 & 64.65 & 80.81 & 82.32 & 83.84 & 85.86 & 87.37 & 80.81 & 84.34 & 90.91 & 90.91 \\
high school government and politics & 80.83 & 88.08 & 71.50 & 91.71 & 89.64 & 90.67 & 92.23 & 92.75 & 87.56 & 95.34 & 94.82 & 95.85 \\
high school macroeconomics & 55.64 & 61.03 & 46.41 & 64.36 & 62.05 & 64.10 & 75.38 & 77.69 & 72.05 & 80.00 & 78.72 & 79.74 \\
high school mathematics & 31.85 & 35.56 & 25.19 & 34.44 & 41.11 & 38.89 & 43.33 & 50.74 & 45.19 & 48.52 & 53.70 & 53.70 \\
high school microeconomics & 62.18 & 73.95 & 59.24 & 73.11 & 72.69 & 72.69 & 84.45 & 88.24 & 82.77 & 88.66 & 88.24 & 88.24 \\
high school physics & 37.75 & 33.11 & 31.13 & 32.45 & 38.41 & 39.07 & 52.32 & 51.66 & 47.02 & 53.64 & 56.95 & 56.29 \\
high school psychology & 81.28 & 83.85 & 74.50 & 83.85 & 83.49 & 84.22 & 88.99 & 89.91 & 84.95 & 91.74 & 90.28 & 90.83 \\
high school statistics & 53.24 & 55.09 & 43.98 & 55.09 & 53.70 & 52.31 & 68.06 & 69.44 & 68.06 & 71.76 & 70.83 & 69.91 \\
high school us history & 75.00 & 77.94 & 68.63 & 79.41 & 77.94 & 77.45 & 85.78 & 86.27 & 81.86 & 87.25 & 85.78 & 87.75 \\
high school world history & 80.17 & 80.17 & 76.79 & 80.17 & 79.32 & 82.28 & 84.39 & 84.81 & 81.86 & 86.50 & 85.65 & 86.08 \\
human aging & 60.09 & 68.61 & 55.16 & 63.23 & 67.71 & 68.61 & 72.20 & 76.23 & 62.33 & 75.34 & 73.99 & 73.09 \\
human sexuality & 70.23 & 76.34 & 65.65 & 77.86 & 77.86 & 77.86 & 76.34 & 77.86 & 75.57 & 81.68 & 82.44 & 82.44 \\
international law & 74.38 & 85.12 & 68.60 & 80.17 & 83.47 & 83.47 & 77.69 & 80.99 & 74.38 & 80.99 & 80.17 & 78.51 \\
jurisprudence & 70.37 & 68.52 & 66.67 & 75.00 & 73.15 & 75.00 & 77.78 & 78.70 & 75.93 & 80.56 & 80.56 & 80.56 \\
logical fallacies & 59.51 & 69.94 & 59.51 & 72.39 & 78.53 & 82.21 & 80.37 & 80.98 & 76.07 & 81.60 & 84.05 & 85.28 \\
machine learning & 33.93 & 42.86 & 43.75 & 44.64 & 43.75 & 41.96 & 47.32 & 46.43 & 47.32 & 47.32 & 49.11 & 48.21 \\
management & 76.70 & 80.58 & 64.08 & 81.55 & 80.58 & 79.61 & 80.58 & 86.41 & 80.58 & 89.32 & 86.41 & 86.41 \\
marketing & 85.04 & 86.32 & 81.62 & 86.75 & 86.32 & 87.61 & 87.61 & 91.45 & 83.76 & 91.88 & 90.60 & 92.31 \\
medical genetics & 75.00 & 78.00 & 69.00 & 75.00 & 78.00 & 82.00 & 81.00 & 80.00 & 77.00 & 79.00 & 78.00 & 80.00 \\
miscellaneous & 75.86 & 77.78 & 71.14 & 78.54 & 79.18 & 80.08 & 83.91 & 85.70 & 81.23 & 85.31 & 85.31 & 85.70 \\
moral disputes & 65.03 & 64.16 & 58.09 & 70.23 & 69.94 & 72.25 & 73.99 & 75.43 & 69.36 & 78.90 & 75.43 & 77.46 \\
moral scenarios & 32.63 & 35.08 & 24.80 & 27.82 & 33.07 & 32.07 & 31.73 & 47.82 & 25.70 & 46.59 & 48.49 & 47.60 \\
nutrition & 70.59 & 73.20 & 61.11 & 75.82 & 75.82 & 77.78 & 79.41 & 80.72 & 75.49 & 80.39 & 80.72 & 80.72 \\
philosophy & 59.81 & 71.38 & 56.91 & 72.99 & 71.38 & 74.60 & 72.67 & 78.46 & 74.60 & 79.10 & 79.42 & 79.74 \\
prehistory & 64.20 & 69.75 & 61.11 & 71.91 & 70.99 & 73.46 & 79.94 & 81.79 & 74.07 & 83.02 & 80.56 & 82.41 \\
professional accounting & 43.97 & 43.97 & 39.72 & 46.45 & 48.94 & 48.94 & 53.19 & 56.03 & 51.06 & 57.45 & 55.67 & 57.45 \\
professional law & 41.85 & 47.13 & 38.59 & 47.59 & 49.93 & 50.20 & 49.09 & 51.04 & 48.04 & 50.98 & 51.83 & 52.09 \\
professional medicine & 69.12 & 72.43 & 45.22 & 75.74 & 72.79 & 72.43 & 77.57 & 77.94 & 74.63 & 75.37 & 73.16 & 74.63 \\
professional psychology & 64.54 & 66.99 & 55.39 & 69.77 & 71.24 & 72.88 & 75.16 & 75.98 & 66.50 & 76.47 & 76.47 & 76.96 \\
public relations & 63.64 & 72.73 & 50.00 & 70.00 & 71.82 & 71.82 & 67.27 & 72.73 & 62.73 & 68.18 & 70.91 & 71.82 \\
security studies & 67.35 & 73.06 & 64.08 & 72.24 & 71.84 & 73.88 & 74.69 & 79.18 & 71.84 & 75.92 & 80.82 & 80.82 \\
sociology & 72.14 & 85.57 & 77.61 & 84.58 & 87.06 & 87.06 & 86.57 & 87.06 & 83.08 & 87.56 & 88.56 & 88.56 \\
us foreign policy & 80.00 & 88.00 & 80.00 & 88.00 & 87.00 & 91.00 & 85.00 & 87.00 & 87.00 & 88.00 & 89.00 & 89.00 \\
virology & 52.41 & 54.22 & 48.19 & 58.43 & 57.23 & 54.82 & 50.00 & 50.60 & 50.00 & 52.41 & 51.81 & 51.81 \\
world religions & 77.19 & 81.87 & 66.08 & 79.53 & 83.04 & 83.04 & 85.96 & 85.38 & 81.29 & 85.38 & 86.55 & 86.55 \\
\midrule
\textbf{All Tasks~(avg)} & 58.00 & 62.18 & 52.29 & 62.65 & 63.79 & \textbf{64.58} & 68.50 & 71.54 & 65.57 & 71.86 & 72.06 & \textbf{72.43} \\
\bottomrule
\end{tabular}
}
\caption{Detailed Results on MMLU for Llama3.1-8B and Qwen2.5-7B. ``ZS'' and ``FS'' denote zero-shot and few-shot prompting, respectively.}
\label{tab:mmlu-all1}
\end{table*}
\begin{table*}[ht]
\centering
\resizebox{\textwidth}{!}{
\begin{tabular}{l|cccccc|cccccc}
\toprule
\multirow{3.5}{*}{\textbf{MMLU Tasks}} & \multicolumn{6}{c|}{\textbf{Mistral-7B}} & \multicolumn{6}{c}{\textbf{GPT-4o-mini}} \\
\cmidrule(lr){2-7} \cmidrule(lr){8-13}
& ZS & FS & Self-ICL & DAIL & \OURS & \begin{tabular}[c]{@{}c@{}}\OURS\\ w/o cache\end{tabular} & ZS & FS & Self-ICL & DAIL & \OURS & \begin{tabular}[c]{@{}c@{}}\OURS\\ w/o cache\end{tabular}  \\
\midrule
abstract algebra & 26.00 & 31.00 & 26.00 & 31.00 & 30.00 & 30.00 & 24.00 & 30.00 & 31.00 & 26.00 & 36.00 & 35.00 \\
anatomy & 56.30 & 57.78 & 49.63 & 58.52 & 60.74 & 62.22 & 53.33 & 62.22 & 61.48 & 44.44 & 53.33 & 56.30 \\
astronomy & 55.26 & 63.16 & 51.32 & 65.79 & 63.82 & 61.84 & 61.84 & 48.03 & 73.68 & 57.89 & 75.66 & 79.61 \\
business ethics & 59.00 & 57.00 & 48.00 & 54.00 & 58.00 & 54.00 & 62.00 & 51.00 & 69.00 & 57.00 & 59.00 & 66.00 \\
clinical knowledge & 67.55 & 67.55 & 53.21 & 70.57 & 72.08 & 72.08 & 50.94 & 67.92 & 66.79 & 55.47 & 64.91 & 62.64 \\
college biology & 66.67 & 67.36 & 59.72 & 68.75 & 68.06 & 70.83 & 52.78 & 67.36 & 72.22 & 70.83 & 80.56 & 79.17 \\
college chemistry & 34.00 & 46.00 & 30.00 & 50.00 & 45.00 & 47.00 & 34.00 & 30.00 & 30.00 & 35.00 & 39.00 & 45.00 \\
college computer science & 46.00 & 44.00 & 38.00 & 45.00 & 49.00 & 47.00 & 40.00 & 42.00 & 45.00 & 47.00 & 52.00 & 61.00 \\
college mathematics & 34.00 & 37.00 & 33.00 & 40.00 & 39.00 & 41.00 & 27.00 & 25.00 & 28.00 & 31.00 & 29.00 & 38.00 \\
college medicine & 58.38 & 60.12 & 49.13 & 62.43 & 62.43 & 63.58 & 39.88 & 50.87 & 53.76 & 52.02 & 60.69 & 58.38 \\
college physics & 37.25 & 37.25 & 32.35 & 34.31 & 34.31 & 36.27 & 33.33 & 35.29 & 35.29 & 39.22 & 43.14 & 52.94 \\
computer security & 75.00 & 79.00 & 70.00 & 82.00 & 78.00 & 79.00 & 62.00 & 65.00 & 68.00 & 60.00 & 66.00 & 67.00 \\
conceptual physics & 47.23 & 54.89 & 44.26 & 51.91 & 55.74 & 55.74 & 51.06 & 40.43 & 56.60 & 45.11 & 47.23 & 48.51 \\
econometrics & 37.72 & 44.74 & 31.58 & 44.74 & 41.23 & 42.98 & 28.95 & 42.11 & 50.88 & 48.25 & 58.77 & 57.89 \\
electrical engineering & 53.79 & 60.00 & 51.03 & 55.17 & 59.31 & 59.31 & 38.62 & 44.14 & 50.34 & 40.00 & 47.59 & 51.72 \\
elementary mathematics & 34.66 & 38.89 & 28.31 & 39.15 & 38.89 & 41.01 & 28.31 & 36.77 & 34.92 & 34.39 & 40.48 & 43.65 \\
formal logic & 38.10 & 36.51 & 36.51 & 38.10 & 41.27 & 37.30 & 27.78 & 30.16 & 38.89 & 39.68 & 34.92 & 32.54 \\
global facts & 37.00 & 34.00 & 20.00 & 34.00 & 38.00 & 33.00 & 39.00 & 27.00 & 37.00 & 28.00 & 29.00 & 28.00 \\
high school biology & 70.97 & 73.55 & 61.94 & 74.52 & 76.13 & 77.74 & 54.19 & 65.48 & 74.52 & 58.06 & 69.68 & 69.68 \\
high school chemistry & 47.29 & 48.28 & 35.47 & 48.77 & 50.25 & 48.77 & 30.05 & 48.28 & 45.32 & 39.41 & 50.74 & 48.77 \\
high school computer science & 63.00 & 65.00 & 55.00 & 65.00 & 64.00 & 67.00 & 44.00 & 70.00 & 53.00 & 57.00 & 75.00 & 74.00 \\
high school european history & 72.12 & 72.73 & 69.70 & 75.76 & 70.30 & 74.55 & 72.12 & 86.06 & 81.82 & 86.06 & 86.06 & 87.27 \\
high school geography & 73.74 & 77.27 & 61.11 & 77.78 & 78.28 & 80.30 & 76.26 & 78.28 & 71.21 & 60.61 & 73.23 & 72.73 \\
high school government and politics & 81.35 & 86.53 & 70.47 & 87.56 & 86.53 & 86.53 & 76.17 & 85.49 & 88.08 & 77.72 & 87.56 & 91.19 \\
high school macroeconomics & 56.15 & 62.82 & 45.64 & 62.82 & 64.36 & 64.62 & 45.90 & 61.79 & 65.64 & 47.44 & 50.26 & 53.08 \\
high school mathematics & 27.78 & 35.93 & 23.70 & 31.11 & 31.48 & 33.70 & 27.78 & 22.96 & 21.11 & 26.30 & 27.78 & 22.59 \\
high school microeconomics & 59.24 & 65.55 & 48.32 & 66.81 & 67.23 & 65.13 & 48.74 & 55.88 & 65.55 & 47.06 & 65.97 & 69.75 \\
high school physics & 34.44 & 39.07 & 22.52 & 34.44 & 40.40 & 43.05 & 32.45 & 47.02 & 41.06 & 42.38 & 53.64 & 49.01 \\
high school psychology & 77.98 & 77.98 & 64.77 & 80.92 & 79.82 & 80.92 & 65.87 & 74.86 & 78.72 & 62.57 & 73.58 & 76.33 \\
high school statistics & 42.59 & 51.39 & 33.33 & 44.44 & 46.30 & 43.98 & 36.57 & 44.44 & 50.93 & 47.22 & 57.41 & 59.26 \\
high school us history & 75.49 & 81.37 & 74.02 & 81.86 & 82.35 & 82.84 & 78.92 & 91.18 & 83.33 & 89.22 & 93.14 & 92.65 \\
high school world history & 74.68 & 75.11 & 73.84 & 75.95 & 75.95 & 75.95 & 74.26 & 89.45 & 86.50 & 89.03 & 88.61 & 89.45 \\
human aging & 66.82 & 69.06 & 62.33 & 69.06 & 68.16 & 70.40 & 66.37 & 64.13 & 61.88 & 57.40 & 69.51 & 64.57 \\
human sexuality & 68.70 & 74.81 & 63.36 & 78.63 & 77.86 & 77.10 & 60.31 & 61.07 & 69.47 & 58.02 & 61.83 & 64.12 \\
international law & 68.60 & 76.03 & 66.94 & 78.51 & 80.17 & 77.69 & 70.25 & 87.60 & 85.95 & 79.34 & 90.08 & 90.08 \\
jurisprudence & 68.52 & 74.07 & 62.04 & 72.22 & 74.07 & 75.00 & 62.96 & 69.44 & 75.93 & 57.41 & 70.37 & 72.22 \\
logical fallacies & 71.17 & 74.85 & 65.03 & 74.23 & 76.07 & 80.98 & 55.21 & 60.12 & 69.33 & 60.12 & 66.87 & 64.42 \\
machine learning & 40.18 & 50.00 & 41.07 & 51.79 & 52.68 & 47.32 & 45.54 & 41.96 & 40.18 & 47.32 & 55.36 & 58.93 \\
management & 66.02 & 77.67 & 64.08 & 78.64 & 79.61 & 78.64 & 73.79 & 75.73 & 71.84 & 60.19 & 66.02 & 71.84 \\
marketing & 82.91 & 86.75 & 79.06 & 88.46 & 86.75 & 88.46 & 73.50 & 88.03 & 79.91 & 79.91 & 85.90 & 89.32 \\
medical genetics & 64.00 & 70.00 & 57.00 & 70.00 & 68.00 & 70.00 & 55.00 & 70.00 & 67.00 & 57.00 & 69.00 & 69.00 \\
miscellaneous & 75.35 & 80.33 & 70.63 & 79.95 & 80.08 & 79.82 & 77.65 & 89.53 & 83.40 & 68.71 & 82.25 & 82.38 \\
moral disputes & 65.03 & 69.65 & 60.69 & 70.23 & 70.52 & 72.54 & 69.08 & 66.47 & 69.36 & 69.94 & 79.19 & 78.90 \\
moral scenarios & 27.93 & 31.84 & 22.57 & 25.47 & 38.55 & 41.79 & 36.09 & 47.15 & 36.31 & 44.25 & 48.49 & 48.27 \\
nutrition & 66.34 & 71.24 & 62.09 & 71.57 & 74.18 & 72.22 & 51.96 & 66.34 & 62.09 & 54.58 & 66.01 & 63.73 \\
philosophy & 65.27 & 70.10 & 61.74 & 71.06 & 69.77 & 71.06 & 55.63 & 62.70 & 64.63 & 67.20 & 73.63 & 75.56 \\
prehistory & 66.67 & 64.81 & 62.35 & 68.52 & 68.83 & 69.44 & 66.98 & 83.64 & 68.52 & 62.96 & 66.98 & 71.91 \\
professional accounting & 46.45 & 43.97 & 38.30 & 48.23 & 46.10 & 45.74 & 38.30 & 52.13 & 49.65 & 51.42 & 57.09 & 58.51 \\
professional law & 44.26 & 44.98 & 38.33 & 45.63 & 46.54 & 46.54 & 43.48 & 59.65 & 52.80 & 59.00 & 59.78 & 60.10 \\
professional medicine & 58.46 & 65.44 & 57.72 & 63.60 & 63.97 & 63.24 & 23.90 & 61.03 & 51.84 & 61.40 & 67.65 & 83.46 \\
professional psychology & 58.33 & 65.52 & 54.41 & 67.32 & 68.46 & 67.65 & 50.16 & 78.43 & 65.85 & 66.01 & 75.98 & 78.76 \\
public relations & 61.82 & 68.18 & 60.00 & 67.27 & 69.09 & 70.91 & 67.27 & 62.73 & 61.82 & 52.73 & 55.45 & 60.91 \\
security studies & 61.22 & 71.02 & 55.92 & 68.16 & 71.02 & 73.88 & 63.67 & 77.14 & 74.69 & 68.57 & 75.92 & 75.51 \\
sociology & 79.60 & 83.08 & 76.12 & 85.57 & 83.58 & 83.58 & 68.16 & 72.64 & 78.61 & 74.63 & 82.09 & 86.07 \\
us foreign policy & 80.00 & 85.00 & 80.00 & 82.00 & 85.00 & 83.00 & 79.00 & 76.00 & 89.00 & 72.00 & 84.00 & 84.00 \\
virology & 48.80 & 51.81 & 46.99 & 49.40 & 51.20 & 52.41 & 43.37 & 51.81 & 43.37 & 40.96 & 42.17 & 47.59 \\
world religions & 80.70 & 81.29 & 77.78 & 80.70 & 80.12 & 83.04 & 76.02 & 80.12 & 74.85 & 67.84 & 80.70 & 82.46 \\
\midrule
\textbf{All Tasks~(avg)} & 57.01 & 60.70 & 51.04 & 60.69 & 61.98 & \textbf{62.54} & 52.28 & 62.60 & 60.88 & 57.22 & 64.26 & \textbf{65.62} \\
\bottomrule
\end{tabular}
}
\caption{Detailed results on MMLU tasks for Mistral-7B and GPT-4o-mini. ``ZS'' and ``FS'' denote zero-shot and few-shot prompting, respectively.}
\label{tab:mmlu-all2}
\end{table*}

\end{document}